\documentclass[lettersize,journal]{IEEEtran}
\usepackage{amsmath,amsfonts}
\usepackage{algorithm}
\usepackage{array}
\usepackage[caption=false,font=normalsize,labelfont=sf,textfont=sf]{subfig}
\usepackage{textcomp}
\usepackage{stfloats}
\usepackage{url}
\usepackage{verbatim}
\usepackage{graphicx}
\usepackage{cite}
\usepackage{graphicx}
\usepackage{booktabs}

\usepackage{hyperref}
\usepackage{wrapfig}

\usepackage{xcolor}
\usepackage{algorithm}
\usepackage{algpseudocode}
\definecolor{revision-blue}{RGB}{0,0,0}

\makeatletter
\newcommand\fs@betterruled{%
  \def\@fs@cfont{\bfseries}\let\@fs@capt\floatc@ruled
  \def\@fs@pre{\vspace*{5pt}\hrule height.8pt depth0pt \kern2pt}%
  \def\@fs@post{\kern2pt\hrule\relax}%
  \def\@fs@mid{\kern2pt\hrule\kern2pt}%
  \let\@fs@iftopcapt\iftrue}
\floatstyle{betterruled}
\restylefloat{algorithm}
\makeatother

\begin{document}

\title{Uncertainty-Aware Deep Video Compression with Ensembles}

\author{Wufei Ma,~\IEEEmembership{Student Member,~IEEE}, Jiahao Li, Bin Li,~\IEEEmembership{Member,~IEEE}, Yan Lu,~\IEEEmembership{Member,~IEEE}
\IEEEcompsocitemizethanks{
\IEEEcompsocthanksitem W. Ma is with Johns Hopkins University, E-mail: wma27@jhu.edu. This work was done when W. Ma was a full-time intern with Microsoft Research Asia.
\IEEEcompsocthanksitem J. Li, B. Li, and Y. Lu are with Microsoft Research Asia, Beijing, China, E-mail: {li.jiahao, libin, yanlu}@microsoft.com}
}



\maketitle

\begin{abstract}
Deep learning-based video compression is a challenging task, and many previous state-of-the-art learning-based video codecs use optical flows to exploit the temporal correlation between successive frames and then compress the residual error. Although these two-stage models are end-to-end optimized, the epistemic uncertainty in the motion estimation and the aleatoric uncertainty from the quantization operation lead to errors in the intermediate representations and introduce artifacts in the reconstructed frames. \textcolor{revision-blue}{This inherent flaw limits the potential for higher bit rate savings. To address this issue, we propose an uncertainty-aware video compression model that can effectively capture the predictive uncertainty with deep ensembles. Additionally, we introduce an ensemble-aware loss to encourage the diversity among ensemble members and investigate the benefits of incorporating adversarial training in the video compression task.} Experimental results on 1080p sequences show that our model can effectively save bits by more than 20\% compared to DVC Pro. 
\end{abstract}


\section{Introduction} \label{sec:introduction}

Video contents are reported to account for 82\% percent of all consumer Internet traffic by 2021, and they are proliferating with an increasing demand for high-resolution videos (e.g., 4K movies) and live streaming services \cite{cisco2020report}. Therefore, we must improve the video compression performance to transmit video with a higher quality given limited Internet bandwidth. In recent years, there has been a surge of deep learning-based video compression models \cite{rippel2019learned,9072487,Agustsson_2020_CVPR,hao2021nerv} and some of them have achieved comparable or even better performance than previous traditional video codecs, such as x264 and x265 \cite{tomar2006converting}.

Although previous deep learning-based video codecs have achieved improved performance on many challenging datasets, most state-of-the-art models estimate deterministic predictions for intermediate representations, such as optical flows and residuals. These models fail to represent the aleatoric uncertainty inherent in the model inputs or the epistemic uncertainty in the model parameters and would blindly assume the predictions to be accurate, which is not always the case \cite{der2009aleatory,kendall2017uncertainties}. In terms of video compression, such models produce deterministic motion vectors (or optical flows) and residuals for each pixel location, ignoring the fact that optical flows may not be estimated accurately in occluded regions and around object boundaries, and the quantization operation before lossless entropy coding also introduces additional noises to the inputs of the decoders. Underlying errors in such overconfident intermediate predictions are propagated to later stages of the P-frame model and even to subsequent frames for models built on temporal correlation, leading to suboptimal performance of the compression system.

Predictive uncertainty is crucial for us to understand how confident the model is about the predictions, especially for out-of-distribution data. However, most neural networks do not offer such information and tend to produce overconfident predictions \cite{Gal2016Uncertainty,NIPS2017_9ef2ed4b}. Bayesian neural networks \cite{mackay1992practical,Hinton1995BayesianLF} are widely used to quantify predictive uncertainty but lack practicality due to significantly increased computation complexity and do not scale well to high-dimensional data. \cite{gal2016dropout} proposed Monte Carlo dropout that performs test-time dropout. It is simple to implement but unsuitable for deep learning-based compression, since it requires multiple decoding-time inferences and yields nondeterministic outputs.

In terms of deep learning-based video compression, two non-Bayesian approaches are considered to represent the predictive uncertainty: (1) modeling the uncertainty explicitly by regressing the empirical variance of the model outputs \cite{374138}; and (2) using ensembles for predictive uncertainty estimation \cite{NIPS2017_9ef2ed4b}. \textcolor{revision-blue}{Scale-space flow \cite{Agustsson_2020_CVPR} took the first approach and proposed to regress a scale field besides the standard 2D flow field, representing the variance associated with each predicted MV (motion vector).} Gaussian blurring is then applied to the reference frame, and the scale parameter is used to control the size of the Gaussian kernel. Although this approach has been shown to be effective, regressed scales are unreliable for out-of-distribution data and are often misinterpreted as predictive uncertainty \cite{Gal2016Uncertainty}.



In this work, we consider the second approach and represent the underlying uncertainty with deep ensembles. Instead of producing a deterministic prediction, ensemble methods perform model combination and reflect the uncertainty of out-of-distribution data. Our ensemble-based decoding module generates an ensemble of intermediate outputs, such as motion vectors and residuals, and implicitly represents the predictive uncertainty with the variance of the Gaussian mixture prediction. This uncertainty is then propagated to later stages, and all modules in our framework are optimized in an end-to-end fashion. Moreover, unlike previous works on whole model-level ensembles, our approach ensembles the partial intermediate layers of the decoding module and achieves improved performance with limited overhead.


To further improve the performance of our uncertainty-aware video compression model, we propose an ensemble-aware loss to encourage diversity between different branches and \textcolor{revision-blue}{incorporate an adversarial training strategy, fast gradient sign method (FGSM) \cite{goodfellow2014explaining}, to effectively learn a smooth latent representation.} Our experiments show that our model can achieve a bitrate saving of more than 20\% on 1080p sequences compared to DVC Pro \cite{9072487}. \textcolor{revision-blue}{Visualizations of the predictive uncertainty captured by our model support our claims and demonstrate the effectiveness of our approach.}

The contributions of this work are summarized as follows:
\begin{itemize}
\item \textcolor{revision-blue}{We identify the underlying uncertainty of intermediate representations as a key limitation of residual-based video compression models and propose an ensemble-based decoder to effectively capture the predictive uncertainty.}
\item \textcolor{revision-blue}{We design a novel ensemble-aware loss to encourage the diversity between ensemble members and better capture the predictive uncertainty. We also show that fast gradient sign method can benefit deep learning-based video compression by learning a smooth intermediate representation.}
\item \textcolor{revision-blue}{Experiments show that our model outperforms previous state-of-the-art models such as DVC Pro \cite{9072487} and scale-space flow \cite{Agustsson_2020_CVPR}, and our approach can be widely applied to optical flow-based video codecs with negligible complexity increase.}
\end{itemize}

\section{Related Work} \label{sec:related-work}

\textbf{Video compression.} Previous learning-based video compression methods can be categorized into two groups: (i) one-stage models, such as methods based on 3D autoencoders \cite{pessoa2020end,habibian2019video}; and (ii) two-stage models, which are adopted by most previous state-of-the-art methods, consist of predicted frame generation and residual coding. \cite{lu2019dvc} proposed an end-to-end trainable video codec, DVC, that utilizes an optical-flow network \cite{Ranjan_2017_CVPR} for motion compensation and then compresses the residuals. DVC Pro \cite{9072487} improves the compression performance by introducing refinement modules and auto-regressive entropy models. \cite{9185043} proposed to learn robust spatio-temporal representations from coding information and to reveal double compression. In order to obtain better motion vectors for motion compensation, \cite{8447515} proposed GPU-based hierarchical motion estimation and \cite{Agustsson_2020_CVPR} proposed scale-space flow to blur intermediate reconstructions when motion vectors are not estimated well. \cite{9681152} designed a cross-resolution synthesis module to pursue better compression efficiency. Moreover, \cite{7736114} exploited the temporal masking effect for better visual qualities.



\textbf{Model uncertainty.} Predictive uncertainty can be grouped into aleatoric uncertainty and epistemic uncertainty \cite{der2009aleatory}. Aleatoric uncertainty captures the noises inherent in the observations and cannot be explained away with more data, while epistemic uncertainty accounts for uncertainty in the model structure or parameters and can be reduced with more training data. Bayesian neural networks \cite{mackay1992practical,Hinton1995BayesianLF} is a widely used approach for modeling predictive uncertainty that extends the traditional neural networks by learning a posterior distribution of model parameters from the observed data. Various non-Bayesian approaches have also been proposed, such as utilizing the probabilities of softmax distributions \cite{hendrycks2016baseline} and Specialists+1 Ensemble for representing the predictive uncertainty for adversarial samples \cite{abbasi2017robustness}. \cite{gal2016dropout} proposed Monte Carlo dropout by performing multiple inferences with dropout at test time. \cite{NIPS2017_9ef2ed4b} proposed to use an ensemble of neural networks for quantifying predictive uncertainty.




\textbf{Deep Ensembles.} The neural networks community has been investigating ensembles of deep networks since the early 1990s \cite{58871,WOLPERT1992241,Perrone1993}. \cite{NIPS1994_b8c37e33} proved the bias-variance trade-off for ensemble models, which suggested the importance of the diversity among ensemble members. \cite{lee2015m} investigated several training strategies to train an ensemble and proposed ensemble-aware oracle loss to encourage diversity. GoogLeNet \cite{7298594}, one of the best-performing models on ILSVRC 2014, is an ensemble of CNNs. \cite{NIPS2017_9ef2ed4b} proposed to estimate predictive uncertainty by training multiple stand-alone neural networks. \cite{NEURIPS2018_be3087e7,fort2020deep} showed that deep ensembles could learn different modes of function with ensemble members that only differ in initialization weights.

\section{Uncertainty-Aware Deep Video Compression} \label{sec:method}

This section presents our main contributions. First, we introduce the theoretical background and the motivation of our proposed approach in Section \ref{sec:method-motivation}. Then we introduce the ensemble-based decoding module to decode multiple candidates of motion vectors and residuals in Section \ref{sec:method-multi-head}. In order to encourage diversity among the ensemble members and to improve the overall performance, we propose an ensemble-aware loss for ensemble-based decoders in Section \ref{sec:method-ensemble}. Finally, we introduce an adversarial training strategy that we find beneficial for the learning-based video compression task in Section \ref{sec:method-fgsm}.

\begin{figure}[!t]
\centering
\subfloat[]{\includegraphics[width=0.8\columnwidth]{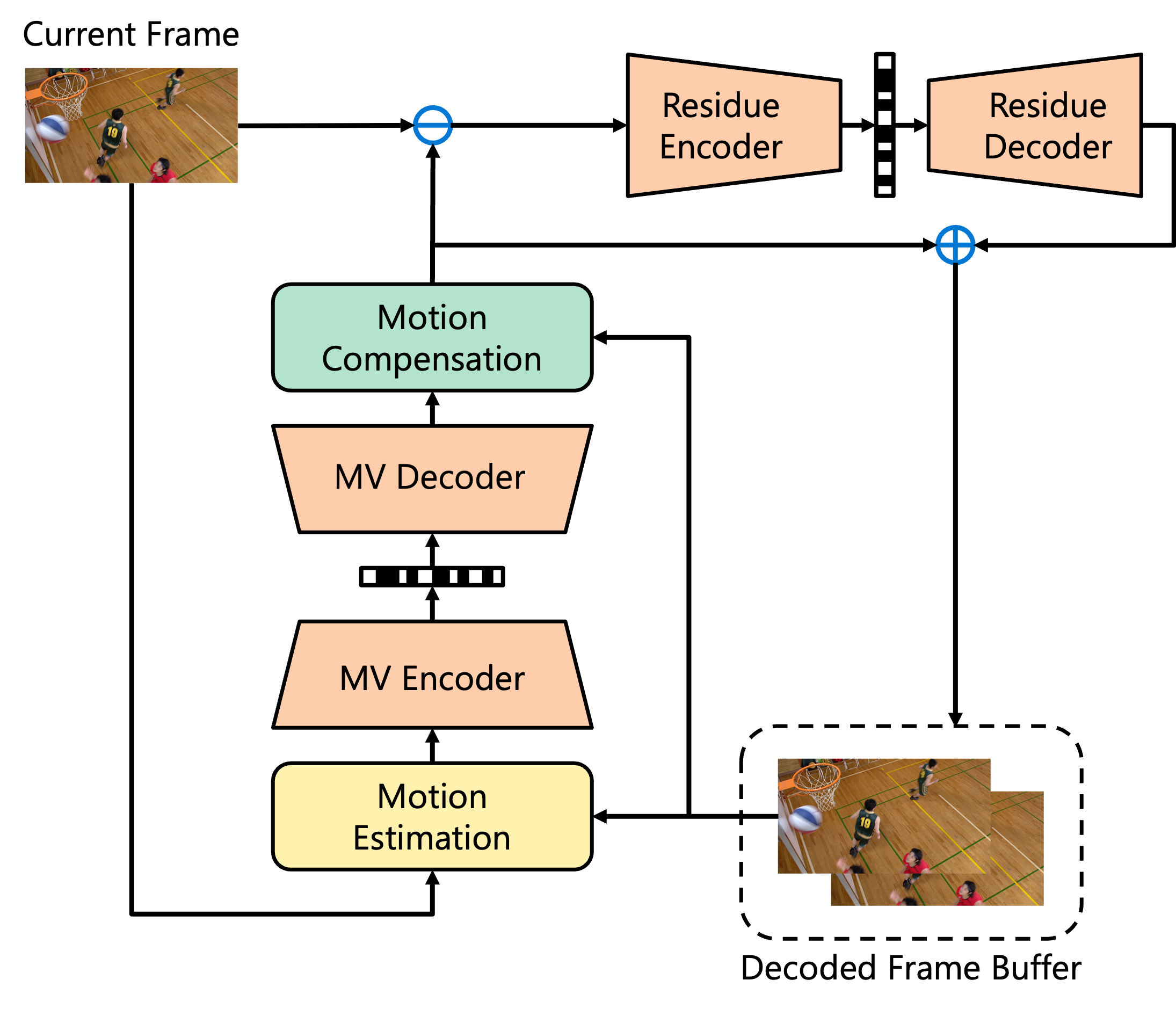}}
\label{fig:fig1-a}
\subfloat[]{\includegraphics[width=0.8\columnwidth]{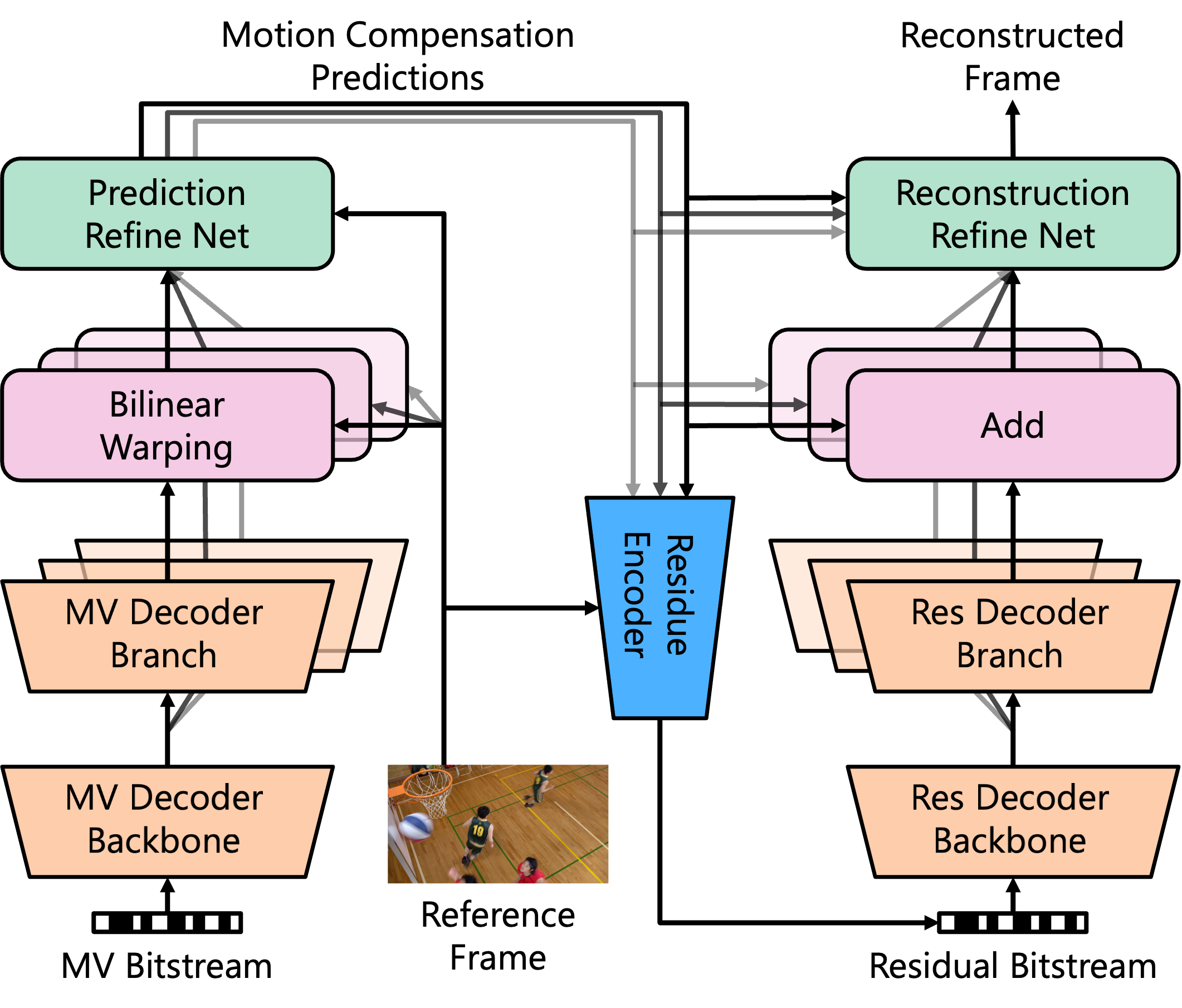}}
\label{fig:fig1-b}
\caption{(a) A low latency predictive coding-based video compression framework. (b) We follow the predictive coding-based video compression and propose ensemble-based decoders.}
\label{fig:fig1}
\end{figure}

\subsection{Uncertainties in Deep Video Compression} \label{sec:method-motivation}

The predictive coding-based model is a popular framework for video compression and is widely used by most previous state-of-the-art models \cite{9072487,lu2020content,hu2020improving}. Let the current frame be $x_t$ and the reconstructed previous frame from the buffer be $\hat{x}_{t-1}$. We estimate a motion vector (MV) map $f_t$ with a motion estimation network. The optical flow is then sent to a motion auto-encoder for transform coding, yielding quantized bits $\hat{a}_t$ and the reconstructed optical flow $\hat{f}_t$. Bilinear warping is used for motion compensation (MC) and an MC prediction $\tilde{x}_t$ with residual $r_t=x_t-\tilde{x}_t$ is obtained. The residual $r_t$ is then compressed with a residual encoder and decoder, outputting quantized residual bitstream $\hat{b}_t$ and the decoded residual $\hat{r}_t$. The reconstructed current frame is the sum of the MC prediction and the decoded residual written as
\begin{align}
    \hat{x}_t = \mathrm{BilinearWarp}(\hat{x}_{t-1}, \hat{f}_t) + \hat{r}_t.
\end{align}


\textbf{Aleatoric uncertainty.} Although at encoding time we have complete information necessary to decode $\hat{f}_t$, for lossy compression at certain bit rates, we quantize the bitstream that is passed to the decoder and inevitably introduces aleatoric uncertainty at decoding time. Since the aleatoric uncertainty cannot be reduced with more training data, a well-trained codec cannot mitigate the quantization noise or fully recover the estimated MV $f_t$.

Consider the MV auto-encoder in the predictive coding-based framework above. The lossy compression of the motion vectors can be summarized as
\begin{align}
    a_t & = \text{MVEncoder}(f_t) \nonumber \\
    \hat{a}_t & = q(a_t) = a_t + \eta \nonumber \\
    \hat{f}_t & = \text{MVDecoder}(\hat{a}_t).
\end{align}
If the MV decoder is implemented with a linear model parameterized by $w$, the impact of the quantization noise $\eta$ on the decoded MV is given by
\begin{align}
w^\top \hat{a}_t = w^\top (a_t + \eta) = w^\top a_t + w^\top\eta.
\end{align}
Since $\eta$ is introduced by the quantization operation, we have $\lVert \eta \rVert_\infty \leq 1/2 = \varepsilon$ and it follows that the upper bound of the effects from the quantization operation is given by
\begin{align}
    \lVert w^\top\eta \rVert_1 \leq \varepsilon \left\lVert w^\top \text{sign}\left( \frac{\partial w^\top a_t}{\partial a_t} \right) \right\rVert_1 = \frac{1}{2} \lVert w^\top \text{sign}( w ) \rVert_1.
\end{align}
While in practice, the MV decoder is usually implemented with a stack of convolution layers and nonlinear activation layers, such as leaky ReLUs, the transformation of the MV decoder may be too linear to reject the quantization noise \cite{goodfellow2014explaining}.

\begin{figure}[!t]
\centering
\subfloat[]{\includegraphics[width=0.48\columnwidth]{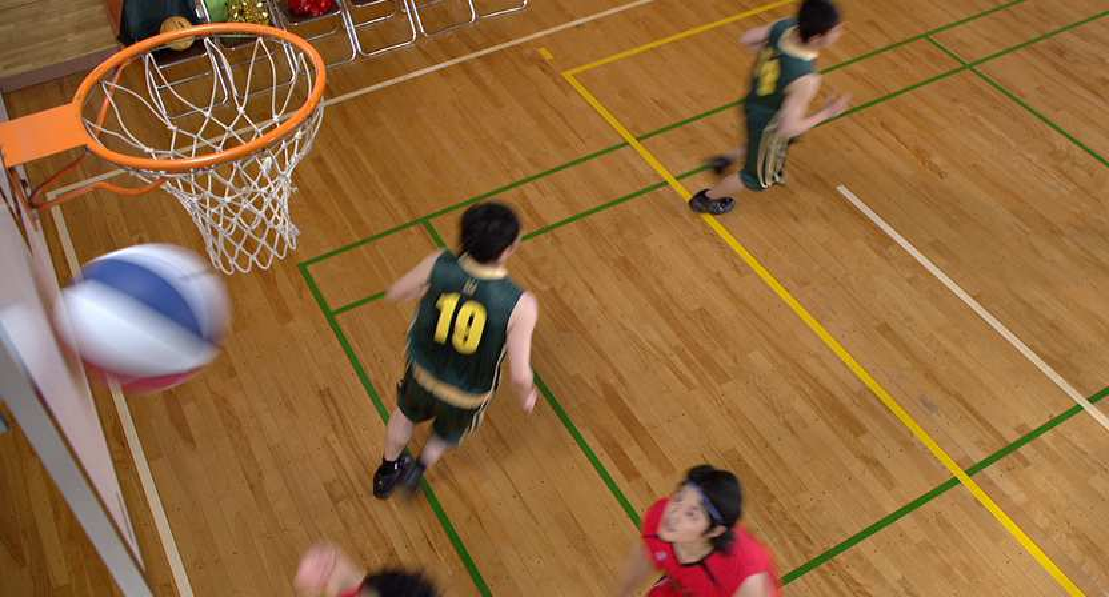}}
\label{fig:fig2-a}
\subfloat[]{\includegraphics[width=0.48\columnwidth]{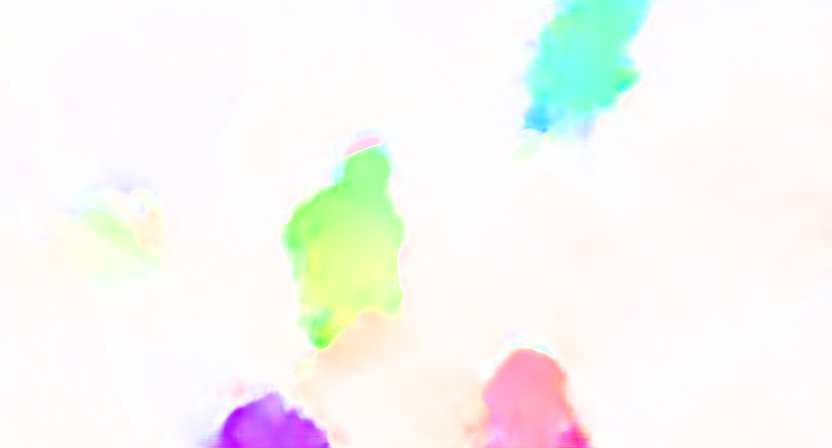}}
\label{fig:fig2-b}
\subfloat[]{\includegraphics[width=0.48\columnwidth]{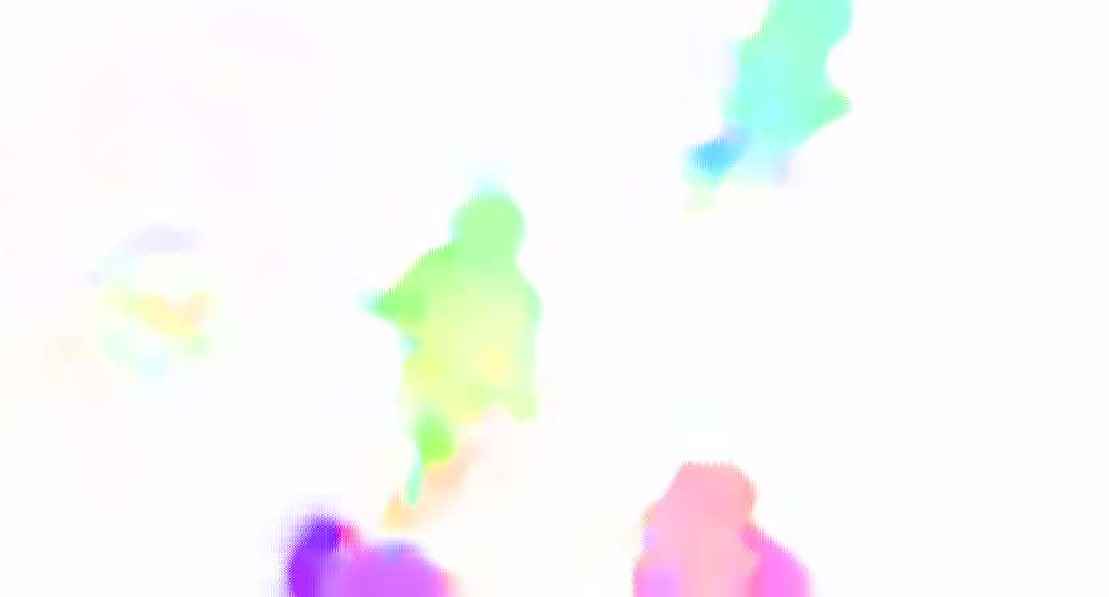}}
\label{fig:fig2-c}
\subfloat[]{\includegraphics[width=0.48\columnwidth]{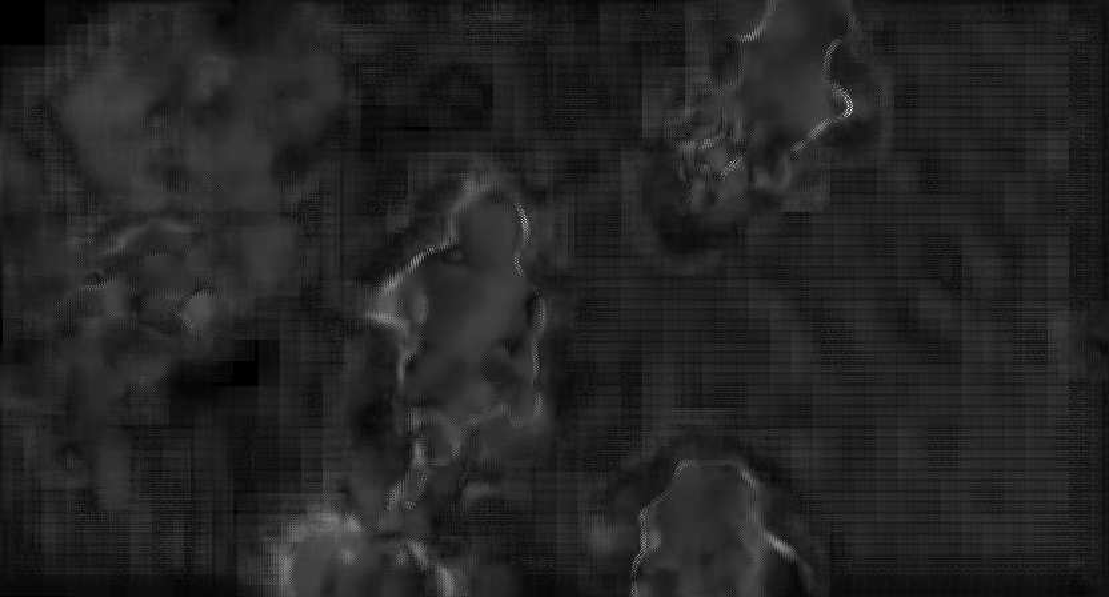}}
\label{fig:fig2-d}
\subfloat[]{\includegraphics[width=0.48\columnwidth]{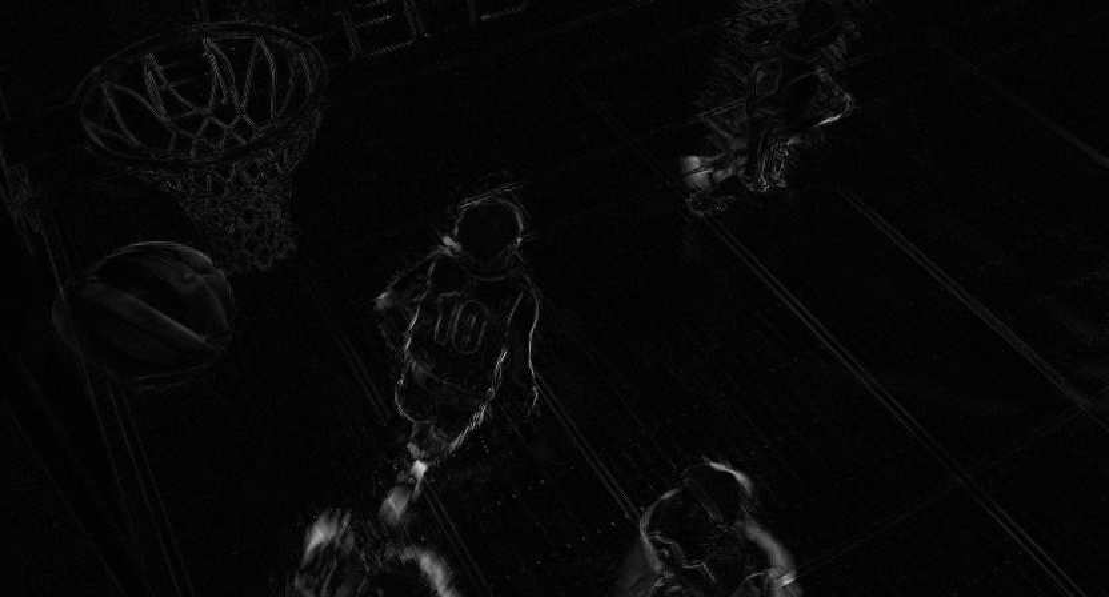}}
\label{fig:fig2-e}
\subfloat[]{\includegraphics[width=0.48\columnwidth]{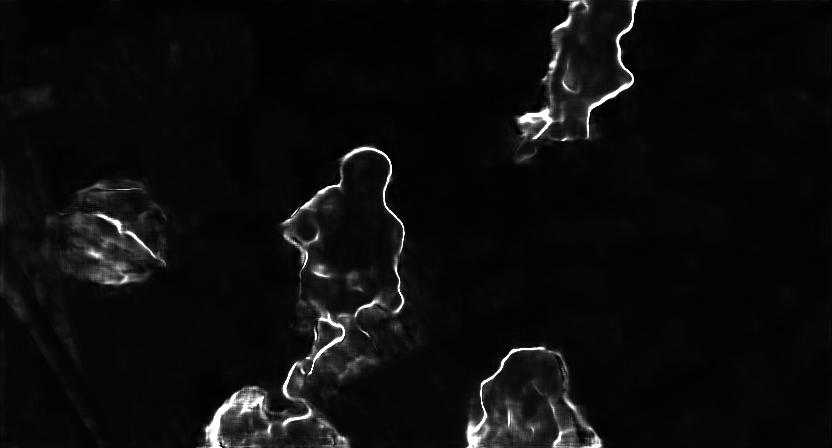}}
\label{fig:fig2-f}
\caption{A preliminary experiment on the underlying uncertainty of the optical flows. (a) The current frame $x_t$ to be compressed. (b) The estimated MV $f_t$. (c) The decoded MV $\hat{f}_t$. (d) Aleatoric uncertainty measured as the L2 distance between two optical flows with and without a small perturbation on the bitstream. (e) Epistemic uncertainty measured by motion vectors that cannot be estimated well. (f) The predictive uncertainty represented by the ensemble-based decoder.}
\label{fig:fig2}
\end{figure}

We conduct preliminary experiments to visualize the aleatoric uncertainty introduced by the quantization operation. We add a small perturbation $\eta_0$ to the quantized bitstream and obtain $\hat{a}_t' = \hat{a}_t + \eta_0$. The perturbation $\eta_0$ is only added to positions where the quantization gap is at least $0.1$ and corresponds to only $20\%$ of the size of the perturbation gap. We model the aleatoric uncertainty with the L1 norm between two optical flows decoded from bitstreams that differs in a small perturbation. Results on the first two frames of the BasketballDrill sequence are shown in Fig. \ref{fig:fig2}(d). As we can see, the aleatoric uncertainty is not uniform across the whole image. Instead, there is more aleatoric uncertainty around the object boundaries and regions where the motion is large. While, by definition, such aleatoric uncertainty cannot be reduced away, blindly assuming the optical flows to be accurate would lead to larger intermediate residual errors and cost more bits in the residual coding.

\textbf{Epistemic uncertainty.} Due to limited observed data during training, epistemic uncertainty accounts for the uncertainty in the model parameters as well as the estimated motion vectors we use to exploit temporal correlation. Motion vectors near the object boundaries and occluded regions tend not to be estimated well, and warping erroneous motion vectors would propagate errors to the residual coding. We may roughly visualize such uncertainty by optimizing a motion estimation network with regards to the mean squared errors (MSE) between the current frame $x_t$ and the warped frame
\begin{align}
    \mathcal{L} = \text{MSE}(x_t, \text{BilinearWarp}(\hat{x}_{t-1}, f_t))
\end{align}
We depict the results in Fig. \ref{fig:fig2}(e). Since the motion estimation is optimized to minimize the MSE, regions where the MSE is large are likely to have a larger epistemic uncertainty and the corresponding motion vectors cannot be estimated well given the limited training data.

Ideally, we could save MV bits by not encoding MVs that are not estimated well and save residual bits by not warping MVs which would not help to reduce residuals. Unfortunately, this is often difficult to implement as an end-to-end optimized deep neural network. In the next section, we will show that with the help of ensemble-based decoders, the model could learn to exploit available information in the bitstream and handle those predictions with larger uncertainty.

\subsection{Ensemble-Based Decoder} \label{sec:method-multi-head}

Our proposed ensemble-based decoder decodes multiple groups of motion vectors (MV) for motion compensation and multiple groups of residuals for the final reconstruction. The ensemble-based MV decoder and residual decoder are depicted in Fig. \ref{fig:fig1}. Take the ensemble-based MV decoder as an example. The MV decoder backbone first decodes a high-dimensional MV feature representation from the quantized MV bitstream $\hat{a}_t$. Then $h$ groups of MVs, denoted by $\{\hat{f}_t^m \mid m = 1, \dots, h\}$, are decoded from the MV feature with respective MV decoder branches. We obtain $h$ warped frames $\{\tilde{x}_t^m \mid m = 1, \dots, h\}$ by bilinearly warping each $\hat{f}_t^m$ on the reference frame $\hat{x}_{t-1}$. The $h$ warped frames are then concatenated for motion compensation and retained for the final reconstruction.


Many previous ensemble-based models train an ensemble of stand-alone neural networks \cite{7298594,abbasi2017robustness,wang2020ensemble}. While they can outperform the \textcolor{revision-blue}{model without ensemble-based decoders} by a wide margin, the number of parameters is greatly increased, as well as the inference complexity. \cite{lee2015m} proposed to share backbone parameters with TreeNets, but the models achieve the best performance when very few layers are shared. In our ensemble-based decoder structure, each decoder branch shares most of the convolution layers, making each decoder branch lightweight. This design effectively improves the overall performance with negligible complexity increase (see Section \ref{sec:exp-ablation}).

The ensemble of decoded MVs can be represented by an equally weighted Gaussian mixture model given by
\begin{align}
    \hat{f}_t \sim \frac{1}{h} \sum_{m=1}^{h} \mathcal{N}(f \mid \hat{f}_t^m, \Sigma_t^m), \; \Sigma_t^m = \begin{bmatrix}
    \sigma_{t,x}^m & 0 \\ 0 & \sigma_{t,y}^m
    \end{bmatrix}
    \label{eq:ours-flow}
\end{align}
where $\sigma_{t,x}^m$ and $\sigma_{t,y}^m$ are the variance in $x$ and $y$ directions respectively. The mean and variance of the Gaussian mixture model are respectively
\begin{align}
    \mathbb{E}[\hat{f}_t] & = \mu_{\hat{f}_t} = \frac{1}{h} \sum_{m=1}^{h} \hat{f}_t^m \\ \sigma_{\hat{f}_{t,x}}^2 & = \frac{1}{h} \sum_{m=1}^{h} \left( \left( \sigma_{t,x}^m \right)^2 + \left( \hat{f}_{t,x}^m \right)^2 \right) - \mu_{\hat{f}_t}^2.
\end{align}

\textbf{How can ensemble-based decoders capture predictive uncertainty?} Our uncertainty-aware model is end-to-end optimized with the rate-distortion loss, but each branch in the ensemble-based decoder is initialized with random weights. With the help of the ensemble-aware loss (Section \ref{sec:method-ensemble}), the functions learned by the decoder branches are diverse in the parameter space but similar in the function space for the training samples. Importantly, for out-of-distribution data in the testing samples, different decoder branches would yield highly varied predictions. We represent this predictive uncertainty as the variance between an ensemble of intermediate representations, and such uncertainty can be propagated between modules (see Fig. \ref{fig:fig1}). After being end-to-end optimized with rate-distortion optimization, each module in our model ``sees'' the predictive uncertainty and learns to process the representation accordingly.

\textbf{Why is predictive uncertainty crucial for learning-based video compression?} Models designed for other vision tasks, such as image recognition or segmentation, often consist of a stack of convolution layers with nonlinear activation functions. Uncertainty in such high-dimensional representations can be easily coded in the magnitude of the values, and noises can be corrected by high-dimensional nonlinear mappings. However, in learning-based video compression, the models are built on 2D optical flows and quantized bitstream. Errors in the optical flows and the quantization noises in the bitstream lead to artifacts in the reconstructed frames. Although we cannot ignore ``bad'' MVs, we can alleviate the influence of such MVs by refining the warped frames with the learned uncertainty information. Similarly, we could relieve the artifacts introduced by the quantization noise by processing an ensemble of decoded residuals from the parallel decoder branches.

\textbf{Visualization of the predictive uncertainty.} In order to investigate the predictive uncertainty learned by the ensemble-based decoders and to confirm that the benefit of ensemble-based decoders is not due to extra model complexity or additional non-linearity (from bilinear warping), we conduct preliminary experiments. Empirically, we visualize the predictive uncertainty represented by this ensemble model with the variance of the Gaussian mixture model by setting $(\sigma_{t,x}^m)^2 = (\sigma_{t,y}^m)^2 = 1$, which gives
\begin{align}
    \sigma_{\hat{f}_{t,x}}^2 = \frac{1}{h} \sum_{m=1}^{h} \left( \hat{f}_{t,x}^m \right)^2 - \left( \frac{1}{h} \sum_{m=1}^{h} \hat{f}_{t,x}^m \right)^2 + 1.
    \label{eq:uncertainty}
\end{align}
The predictive uncertainty for the first two frames in the BasketballDrill sequence is depicted in Fig.~\ref{fig:fig2}. \textcolor{revision-blue}{Results from more video sequences are shown in Fig.~\ref{fig:fig-5}.} We can see that the predictive uncertainty estimated by the ensemble-based decoders can properly capture both the aleatoric uncertainty and the epistemic uncertainty shown in Fig. \ref{fig:fig2} --- the basketball and human body parts have large aleatoric uncertainty due to rapid motion and the object boundaries have large epistemic uncertainty.

\textbf{Relation to scale-space flow.} \cite{Agustsson_2020_CVPR} estimated a scale field $\hat{\sigma}_t$ besides the 2-dimensional optical flow $(\hat{f}_{t,x}, \hat{f}_{t,y})$. We may represent the decoded MV with a multivariate Gaussian distribution given by
\begin{align}
    \hat{f}_t \sim \mathcal{N}((\hat{f}_{t,x}, \hat{f}_{t,y})^\top, \Sigma), \; \Sigma = \begin{bmatrix}
    \hat{\sigma}_t & 0 \\ 0 & \hat{\sigma}_t
    \end{bmatrix}
    \label{eq:ss-flow}
\end{align}
and the scale-space warp gives a weighted mean of the warped value obtained from $\hat{f}_t$. As we can see, the MV prediction from our proposed ensemble-based decoder (Eq. \ref{eq:ours-flow}) can represent a more diverse distribution than the multivariate Gaussian distribution from the scale-space flow (Eq. \ref{eq:ss-flow}). On the other hand, the regressed variance $\hat{\sigma}_t$ can be unreliable for out-of-distribution data and is often mis-interpreted as the predictive uncertainty \cite{Gal2016Uncertainty}. Instead, ensemble models produce diverse results by learning different modes of the function, rather than interpolating around a given mean in the output space \cite{NEURIPS2018_be3087e7,fort2019deep}. Given out-of-distribution inputs, each decoder branch would perform very differently and our ensemble-based decoder can capture the predictive uncertainty from the Gaussian mixture representation.

\subsection{Ensemble-Aware Training} \label{sec:method-ensemble}

Intuitively, diversity is a key factor for ensemble models.  Ensemble members similar in the parameter space are unlikely to provide any more useful information than their \textcolor{revision-blue}{non-ensemble counterparts}. \cite{NIPS1994_b8c37e33} proved the bias-variance trade-off in ensemble, $E = \bar{E} - \bar{A}$, which suggested that the inherent variance is the key for the ensemble models to be effective and we should encourage the diversity among the ensemble members.


In the previous literature, multiple approaches are considered, including random initialization, bagging, and boosting. Randomly initializing the model parameters is a simple but effective approach to induce randomness and is quite suitable for deep ensembles \cite{lee2015m}. Bagging trains ensemble members on independently drawn examples with bootstrap sampling but could harm the model performance since each model may see only 63\% of the available data \cite{lee2015m} and would perform poorly when there is a high correlation inherent in the data \cite{bartlett1998boosting}. Boosting generates the ensemble models sequentially and can be very time consuming for training deep ensembles.

To induce diversity in different branches of our ensemble-based decoding module, we choose to randomly initialize the network parameters, and initial experiments show the efficacy of our approach. To further encourage the diversity among the ensemble members, we propose an ensemble-aware loss that can be applied to any deep ensemble model and induce additional randomness.

Consider a deep ensemble model with $h$ ensemble members and the task is to regress an image $x$. Let the $h$ predictions from the $h$ ensemble members be $\hat{x}^m$ for $m = 1, \dots, h$. \textcolor{revision-blue}{For each 2D location $(i, j)$, let $p$ be the decoder with the $k$-th smallest loss. The ensemble-aware loss is given by}
{\color{revision-blue}\begin{align}
    & \mathcal{L}_\text{ensemble-aware}(x, \hat{x}^1, \dots, \hat{x}^h) = \label{eq:ensemble-loss} \\
    & \sum_{m=1}^h \frac{1}{H\times W} \sum_{1 \leq i, j \leq H,W} \min (\lVert \hat{x}_{i,j}^m - x_{i,j} \rVert_2^2, \lVert \hat{x}_{i,j}^p - x_{i,j} \rVert_2^2) \nonumber
\end{align}}
\textcolor{revision-blue}{where $i,j$ traverses all locations in the 2D lattice. For each 2D location $(i, j)$, the gradient derived from the ensemble-aware loss with respect to the $m$-th ensemble decoder is equivalent to the ensemble member with the $k$-th smallest MSE. As demonstrated in Algorithm~\ref{algo-1}, this loss function can be implemented by clipping the gradients with respect to each ensemble member.}

\begin{algorithm}[t]
\caption{Training with ensemble-aware loss.}
\color{revision-blue}
\label{algo-1}
\begin{algorithmic}[1]
    \State Given reconstructed frame $\hat{x}^1, \dots, \hat{x}^h$.
    \State Compute MSE loss for each reconstruction $\mathcal{L}_\text{MSE}^m \in \mathbb{R}^{H \times W}$  for $m = 1 \dots h$.
    \State Concatenate $\mathcal{L}_\text{MSE}^m$ for $m=1 \dots h$ and obtain a loss matrix $\mathcal{L}_\text{MSE} \in \mathbb{R}^{H \times W \times h}$.
    \For {position $(i,j)$ in 2D lattice}
        \State Let $p$ ($1 \leq p \leq h$) be the decoder with the $k$-th smallest loss in $\mathcal{L}_\text{MSE}^m(i, j)$ for $m = 1 \dots h$.
        \For {$m = 1 \dots h$}
            \State $\mathcal{L}_\text{MSE}^m(i, j) := min(\mathcal{L}_\text{MSE}^m(i, j), \mathcal{L}_\text{MSE}^p(i, j))$.
        \EndFor
    \EndFor
    \State By back propagating $\mathcal{L}^m_\text{MSE}$, the gradients w.r.t. each ensemble decoder is properly clipped.
\end{algorithmic}
\end{algorithm}

The advantage of our ensemble-aware loss is two-fold. On the one hand, this ensemble-aware loss can effectively encourage diversity among the ensemble members.
\textcolor{revision-blue}{With the standard loss, each decoder is forced to perfectly reconstruct every frame regardless of the outputs from other decoders. This would push all decoders to a comparable representation space and perform similarly.}
\textcolor{revision-blue}{Instead, with the ensemble-aware loss, we clip the gradients with respect to decoders with large MSE losses, allowing disagreement between ensemble members.}
\textcolor{revision-blue}{Hence each decoder only aims to perfectly reconstruct a subset of frames, allowing them to explore a more diverse parameter space and as a whole, better capture the predictive uncertainty.} On the other hand, each ensemble member is supervised by all the training samples. The oracle set loss proposed in \cite{lee2015m} assigned exclusive training samples to each ensemble member and significantly harmed the performance of individual ensemble members since each branch only sees a small portion of all training data. Instead, our ensemble-aware loss can effectively encourage diversity among ensemble members, and at the same time, guarantee reliable performance for each ensemble member.

\subsection{Adversarial Training with FGSM}\label{sec:method-fgsm}

Adversarial examples \cite{szegedy2013intriguing} are training samples with small but non-random perturbations that are misclassified by neural networks with high confidence. \cite{goodfellow2014explaining} proposed the fast gradient sign method (FGSM) that applies linear but intentionally worst-case perturbation to the training samples, as given by
\begin{align}
    \eta = \epsilon \cdot \text{sign}(\nabla x J(\theta, x, y))
\end{align}
where $J(\theta, x, y)$ is the cost function, and $\epsilon$ controls the norm of the perturbation. This adversarial training strategy has been shown to boost the image classification performance and improve the model's robustness to adversarial examples. \cite{NIPS2017_9ef2ed4b} interpreted FGSM as an efficient solution to smooth the predictive distributions by increasing the likelihood of the target around an $\epsilon$-neighborhood of the observed training samples.

We find adversarial training with FGSM closely related to learned lossy compression and an effective approach to improve the performance of learned video codecs. In transform coding, we want the latent representation to be as smooth as possible, since after quantization, all latent representations in the $\epsilon$-neighborhood, $\{\hat{a} + \eta \mid \lVert \eta \rVert_\infty < \epsilon\}$, correspond to the same decoded output. Learning a smooth latent representation would help to make the output more robust to quantization noise. Although this could be a natural result of an end-to-end optimized video codec, the experimental results show that FGSM can effectively improve the rate-distortion performance. \textcolor{revision-blue}{Algorithm~\ref{algo-2} summarizes our training pipeline with ensemble-aware training and FGSM.}

\begin{algorithm}[t]
\caption{\textcolor{black}{Overview of our training pipeline.}}
\color{black}
\label{algo-2}
\begin{algorithmic}[1]
    \State Compute $\mathcal{L}_\text{MSE}$ given reconstructed frame $\hat{x}^1, \dots, \hat{x}^h$ and the ground truth frame $x$.
    \State Compute FGSM perturbations $\eta$ based on the gradients w.r.t. $x$: $\eta = \epsilon \times \text{sign}(\nabla_x \mathcal{L}_\text{MSE})$.
    \State Add FGSM perturbations to the ground truth frame $x := x + \eta$, and train network with the ensemble-aware training in Algorithm~\ref{algo-1}.
\end{algorithmic}
\end{algorithm}


\section{Experiments} \label{sec:exp}

\subsection{Experimental Setup} \label{sec:exp-setup}

\textbf{Model architecture.} Our base model architecture follows the design in \cite{9072487}, and we use auto-regressive and hierarchical priors for both the motion vector and residual compression. In order to optimize the model in an end-to-end manner, we need to relax the bits estimation since quantizing the latent bits would make the gradients zero almost everywhere. Following \cite{balle2016end}, we substitute the quantization operation with additive uniform noise during training and perform actual quantization during inference.


\textbf{Training datasets.} Our model is trained on 64,612 video sequences from the training part in Vimeo-90K settuplet dataset \cite{xue2019video}. Each video clip has seven frames with a resolution of $448 \times 256$. We randomly crop the video sequences into $256 \times 256$ pixels during training. Given two successive frames from a random sequence, we treat the first frame as the reference frame, and our model is trained to minimize the rate-distortion cost of encoding and decoding the second frame.

\textbf{Implementation of ensemble-based Decoders.} As depicted in Fig. \ref{fig:fig1}, ensemble-based decoders consist of a shared feature backbone and multiple parallel decoder branches. The decoder branches are lightweight and include two convolution layers with one leaky ReLU in between. For the ensemble-based MV decoder, the decoder backbone first decodes MV feature representation from the MV bitstream $\hat{a}_t$, and then $h$ MVs are decoded with respective MV decoder branches. From the $h$ decoded MVs and the previous decoded frame, we obtain $h$ motion compensation (MC) predictions with bilinear warping. The $h$ MC predictions are concatenated with the previous decoded frame and sent to the Prediction Refine Net, from which we get $h$ refined MC predictions. For the ensemble-based residual decoder, the decoder backbone first decodes the residual feature representation from the residual bitstream $\hat{b}_t$, and then $h$ residuals are decoded with the respective residual decoder branches. From the $h$ decoded residuals and the $h$ refined MC predictions, we obtain $h$ reconstructions. Finally, the $h$ reconstructions are concatenated with $h$ refined MC predictions and sent to the Reconstruction Refine Net, from which we get one refined reconstruction as the final decoded frame. All modules in our model, including decoder backbone, decoder branch, and refine nets, are implemented with neural networks and optimized in an end-to-end manner.

\textbf{Training details.} In our experiments, we adopt the progressive training strategy and warm up the inter-coding module for 150,000 steps with the ensemble-aware motion compensation loss in Eq. \ref{eq:ensemble-loss}. Then the model is end-to-end optimized with the rate-distortion loss given by
\begin{align}
    \mathcal{L}_{RD} = \left( R_\text{mv}(\hat{a}_t) + R_\text{res}(\hat{b}_t) \right) + \lambda \cdot D(x_t, \hat{x}_t)
\end{align}
where $R_\text{mv}(\hat{a}_t)$ and $R_\text{res}(\hat{b}_t)$ represent the numbers of bits used to encode the motion vectors and the residual, $D(F_t, \hat{F}_t)$ measures the distortion in mean squared error or multi-scale structural similarity \cite{wang2003multiscale}, and $\lambda$ is the hyperparameter controlling the trade-off. Four models are trained with different quality rates by setting $\lambda=256, \; 512, \; 1024, \; 2048$. We use the AdamW optimizer \cite{loshchilov2017decoupled} with an initial learning rate of $1\times 10^{-4}$, which is then decreased to $1\times10^{-5}$ after $1.2\times10^6$ steps. Each model is trained on one NVIDIA V100 GPU. \textcolor{revision-blue}{For the main results reported in Section~\ref{sec:exp-quantitative} and Section~\ref{sec:exp-ablation}, we used $h=4$ ensemble decoders, with $k=1$ in ensemble-aware loss and $\epsilon=4/255$ in FGSM.}

\subsection{Quantitative Results} \label{sec:exp-quantitative}

\textbf{Testing datasets.} To show the effectiveness of our proposed uncertainty-aware model, we test our model on the first 100 frames from video sequences in HEVC \cite{sullivan2012overview} with GoP size 10, and the first 120 frames from sequences in UVG \cite{mercat2020uvg}, MCL-JCV \cite{wang2016mcl} with GoP size 12. To balance the trade-off between complexity and performance, we use ensemble-based decoders with $h=4$ members.

\textcolor{revision-blue}{\textbf{Baseline models.} DVC \cite{lu2019dvc} is the pioneer model in deep video compression area. It adopts the residual coding-based framework which is the most common framework in traditional coding standards. DVC Pro \cite{9072487} is the enhanced model of DVC and was also the state-of-the-art model when we developed our algorithm. Thus, considering the significant influence of DVC and DVC Pro, we chose these two models as the benchmarks and tested our algorithms based on DVC Pro.}
%
%
\textcolor{revision-blue}{In order to build the best learning-based video codec, we adopt the state-of-the-art image compression model \cite{Cheng_2020_CVPR} for intra-frame coding. To fairly compare performance and to demonstrate the effectiveness of our approach, we test DVC \cite{lu2019dvc} and DVC Pro \cite{9072487} trained with identical experimental settings and the same intra-frame model, denoted as DVC (cheng2020) and DVC Pro (cheng2020).}

\textcolor{revision-blue}{Moreover, we also compare our methods with other state-of-the-art methods in Table \ref{tab:quant-res} to further exhibit the advantage of our approach. HU\_ECCV20 \cite{hu2020improving} and LU\_ECCV20 \cite{lu2020content} were parallel works based on DVC and addressed the resolution issues and content domains. NeRV\_NeurIPS21 \cite{hao2021nerv} proposed a general neural representation for videos and achieved good performance for video compression by transmitting the model weights for each video.}

\begin{figure*}[!t]
\centering
\subfloat[]{\includegraphics[width=0.32\textwidth]{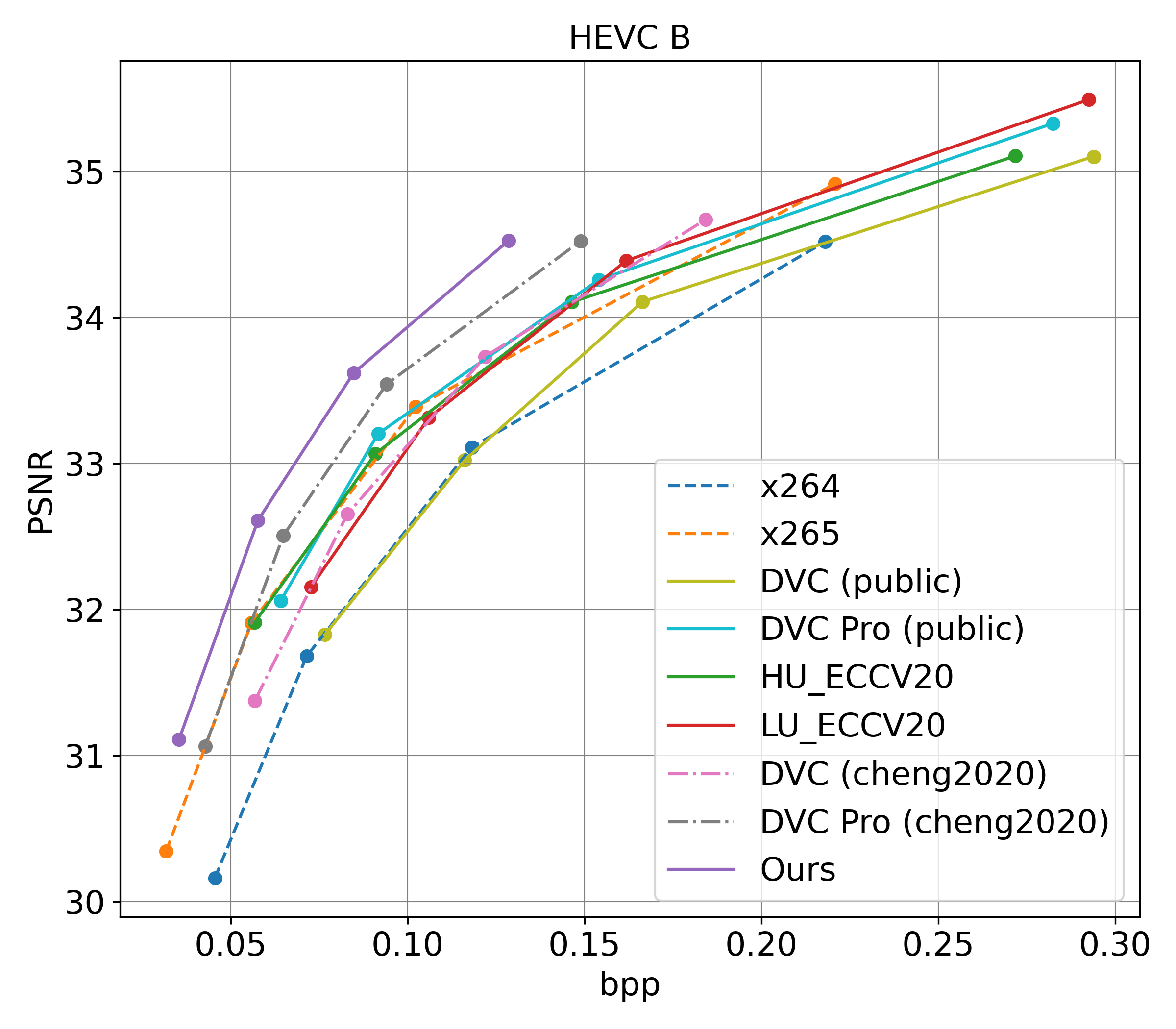}}
\label{fig:fig3-a}
\subfloat[]{\includegraphics[width=0.32\textwidth]{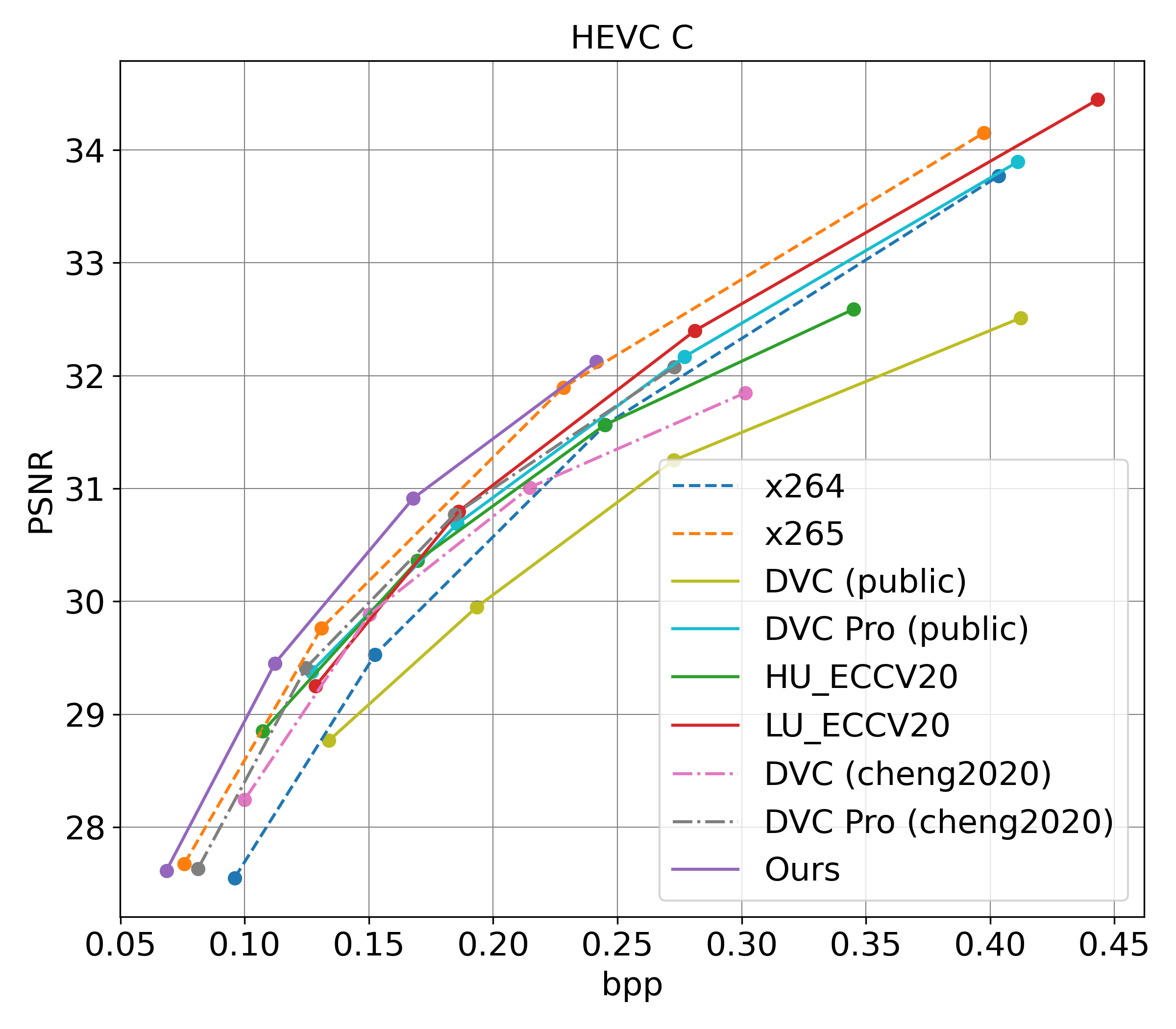}}
\label{fig:fig3-b}
\subfloat[]{\includegraphics[width=0.32\textwidth]{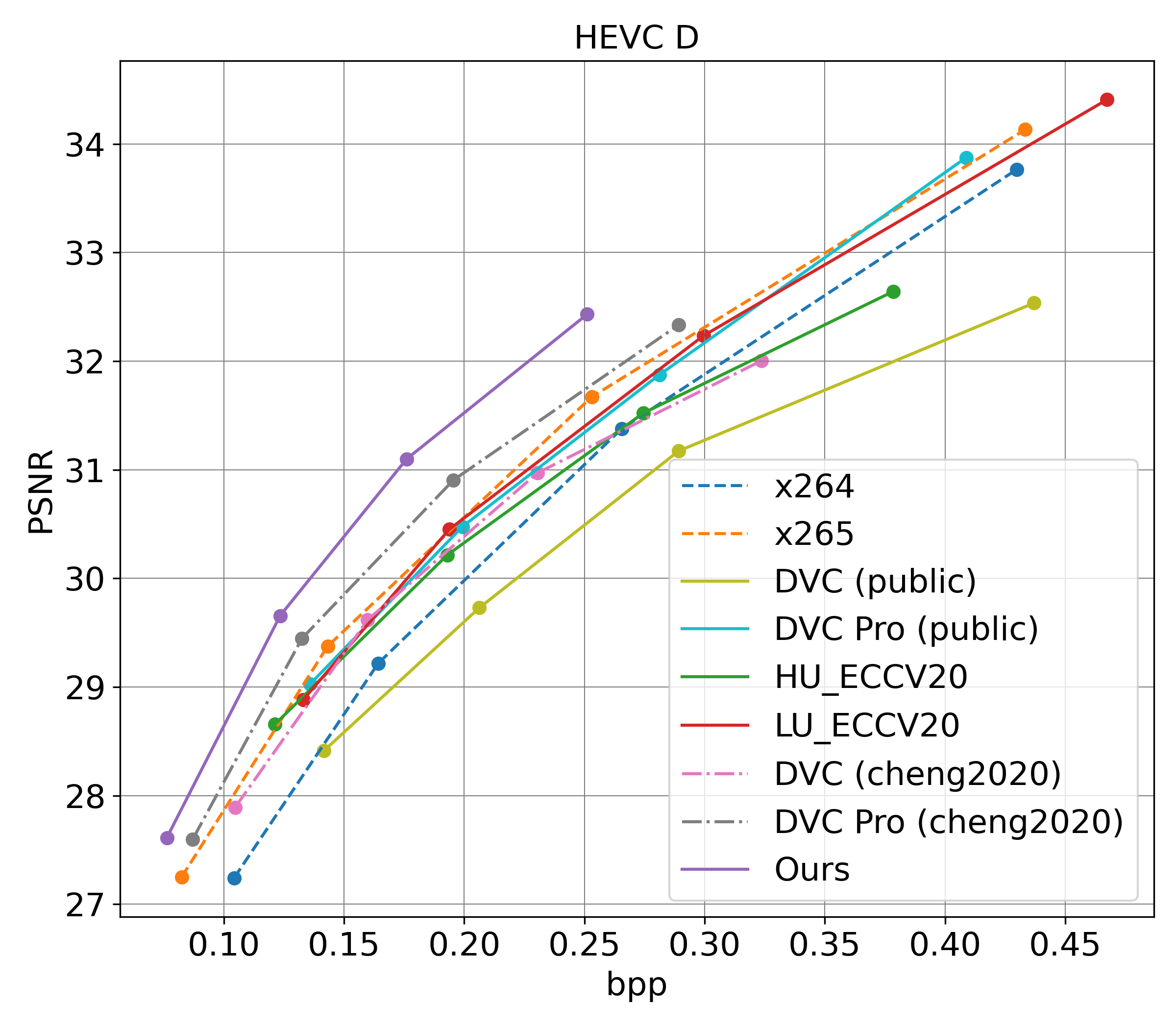}}
\label{fig:fig3-c}
\subfloat[]{\includegraphics[width=0.32\textwidth]{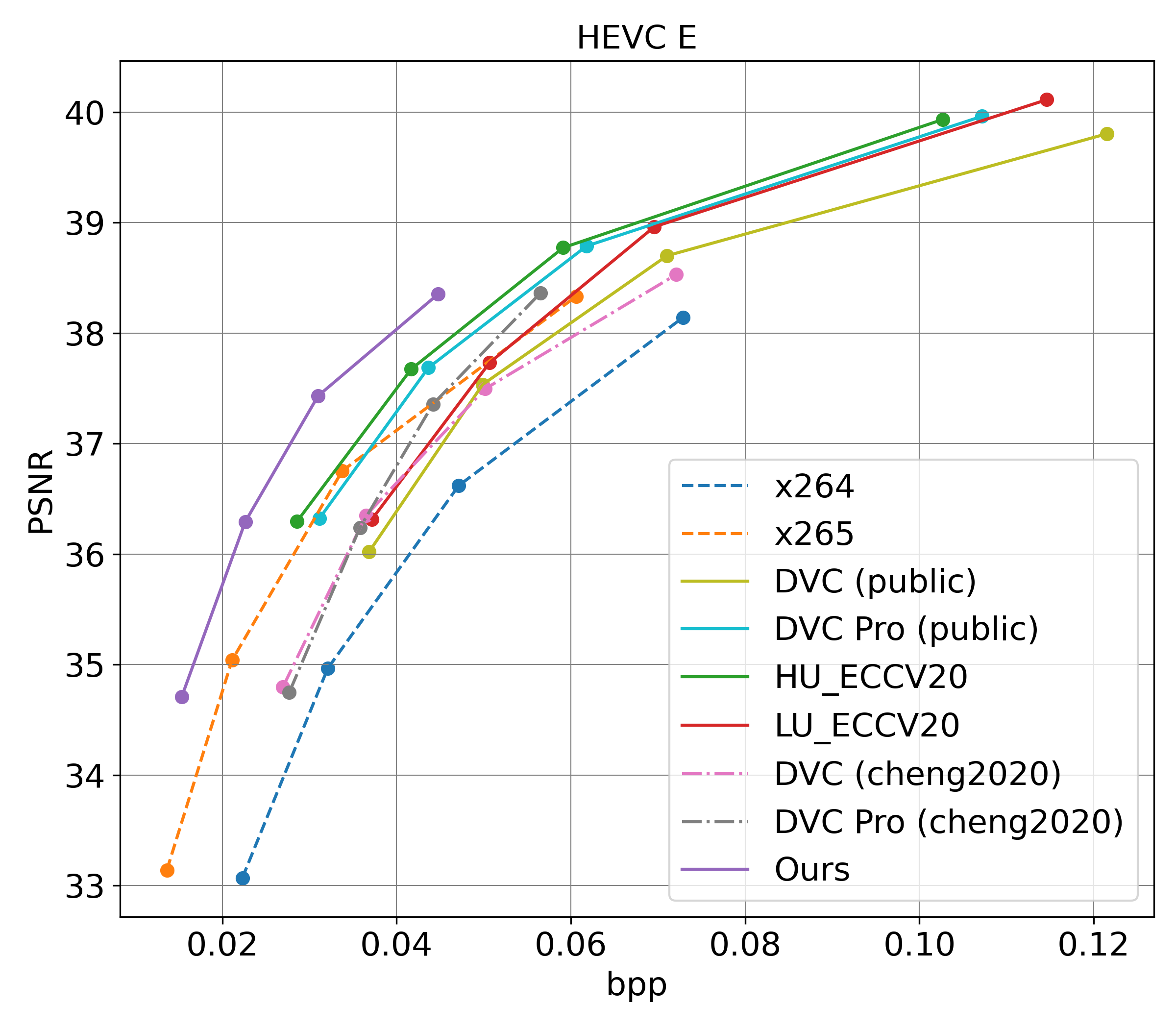}}
\label{fig:fig3-d}
\subfloat[]{\includegraphics[width=0.32\textwidth]{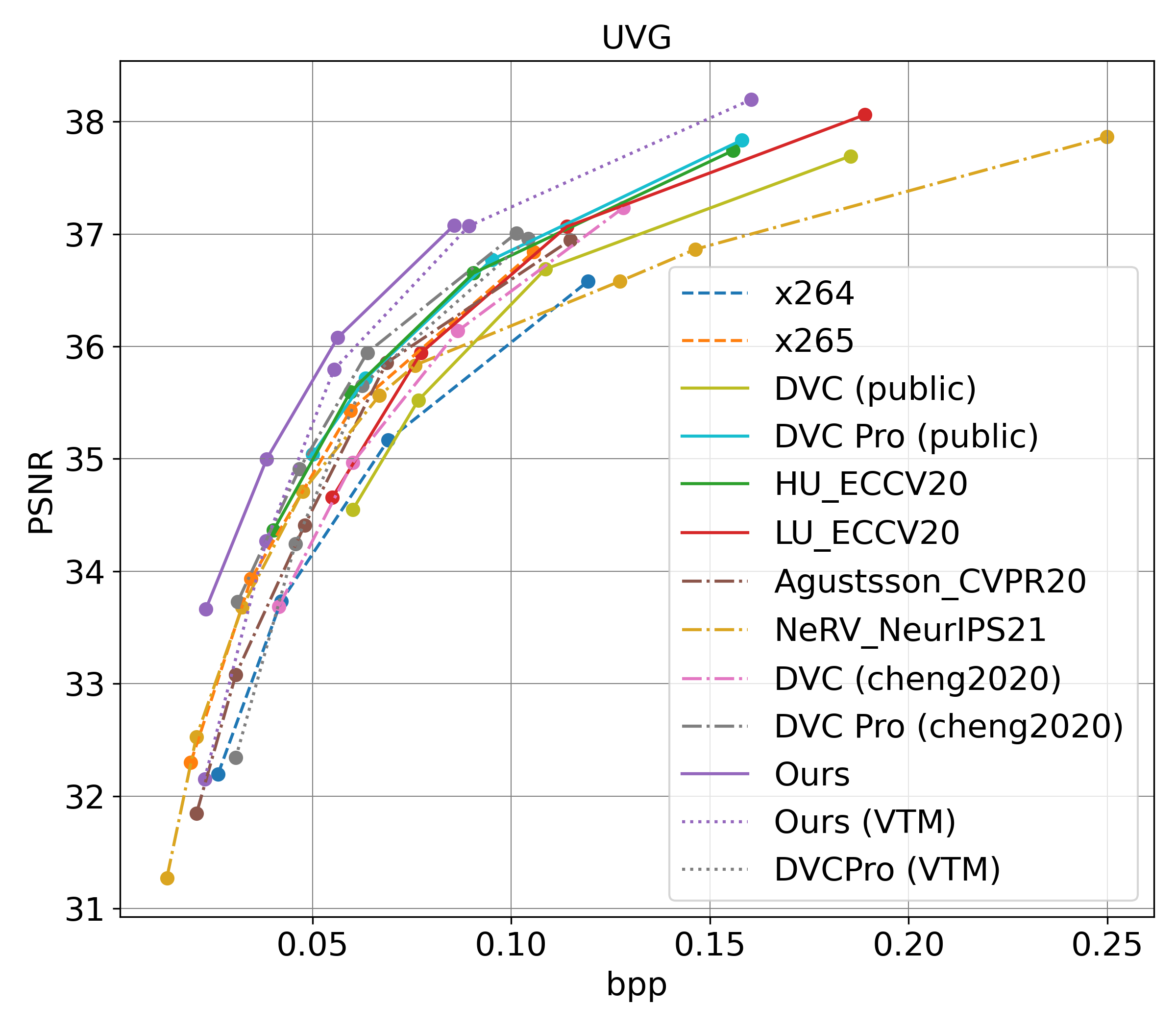}}
\label{fig:fig3-e}
\subfloat[]{\includegraphics[width=0.32\textwidth]{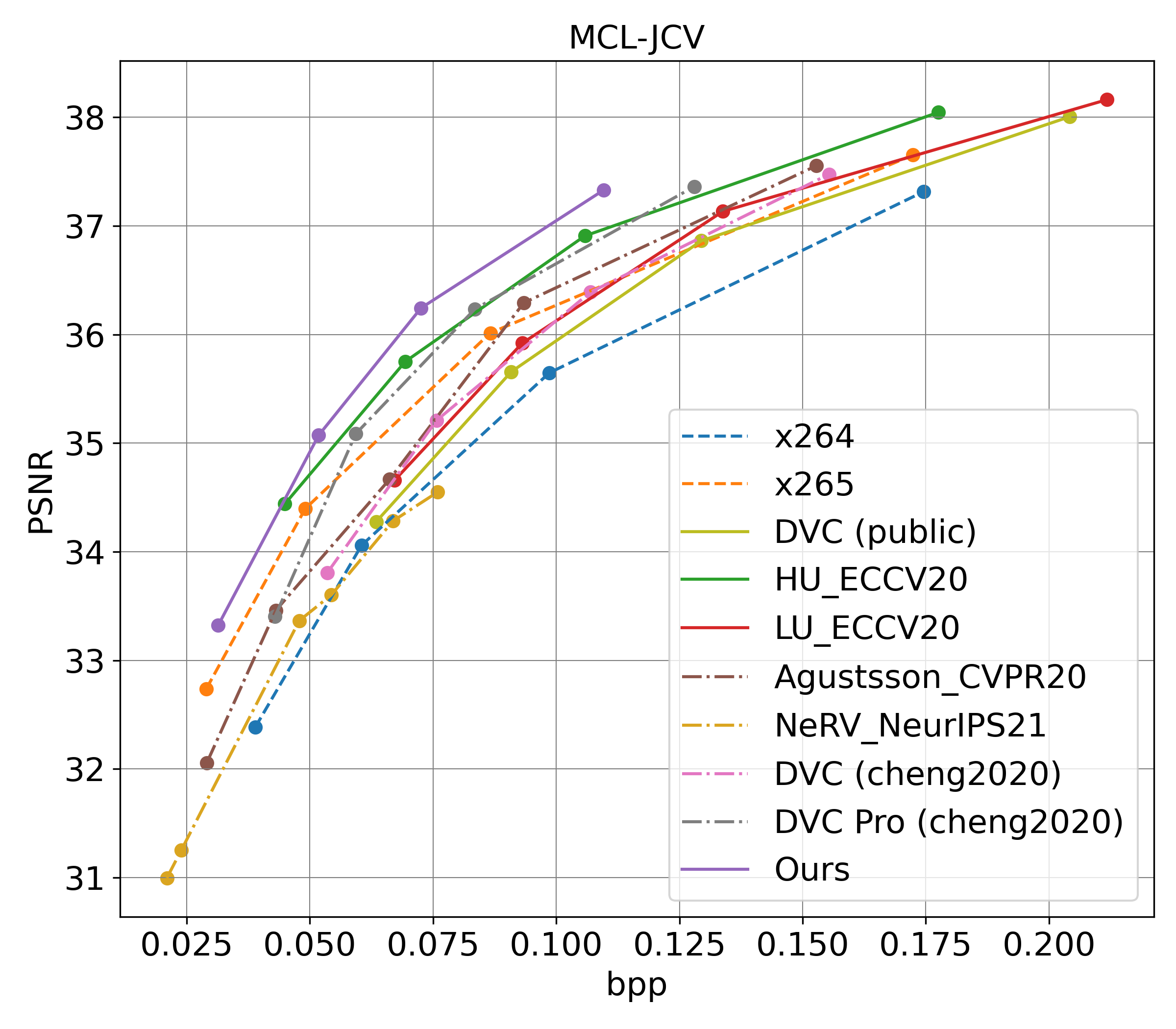}}
\label{fig:fig3-f}
\caption{Rate-distortion comparisons between our model and x264, x265, DVC \cite{lu2019dvc} , DVC Pro \cite{9072487}, Hu\_ECCV20 \cite{hu2020improving}, LU\_ECCV20 \cite{lu2020content}, Agustsson\_CVPR20 \cite{Agustsson_2020_CVPR}, and NeRV \cite{hao2021nerv} on different datasets. \textit{veryslow} preset is used for both x264 and x265. Best viewed in color.}
\label{fig:fig3}
\end{figure*}

We calculate the BD-rate \cite{Bjntegaard2001CalculationOA} of different learning-based video compression models using x.265 as the anchor model, and the results on HEVC, UVG, MCL-JCV are reported in Table \ref{tab:quant-res}. We also plot the RD curves of different codec models in Fig. \ref{fig:fig3}.

\textbf{Quantiative results.} From the results reported in Table \ref{tab:quant-res} and Fig. \ref{fig:fig3}, we could see that our proposed model can effectively save bits compared to our strong baseline and outperform previous state-of-the-art learning-based video codecs by a wide margin in all testing datasets.

\begin{table*}[t]
\caption{Comparisons between different learning-based video compression models measured in BD rate. The anchor model is x265. \textit{veryslow} preset is used for both x264 and x265. A negative number means bitrate saving, and a positive number means bitrate increase.}
\label{tab:quant-res}
\vskip 0.15in
\begin{center}
\begin{tabular}{lcccccc}
\toprule
\multicolumn{1}{c}{\bf MODEL} &\multicolumn{1}{c}{\bf HEVC B} &\multicolumn{1}{c}{\bf HEVC C} &\multicolumn{1}{c}{\bf HEVC D} &\multicolumn{1}{c}{\bf HEVC E} &\multicolumn{1}{c}{\bf UVG} &\multicolumn{1}{c}{\bf MCL-JCV} \\
\midrule
x264  & 35.0\% & 19.9\% & 15.5\% & 50.0\% & 32.7\% & 30.3\% \\
x265  & 0.0\% & 0.0\% & 0.0\% & 0.0\% & 0.0\% & 0.0\% \\
DVC (public) & 26.7\% & 41.5\% & 31.1\% & 17.8\% & 21.9\% & 16.0\% \\
DVC Pro (public) & -0.4\% & 11.5\% & 4.5\% & -3.8\% & -12.6\% & -5.3\% \\
DVC (cheng2020) & 7.9\% & 15.1\% & 7.2\% & 21.1\% & 17.2\% & 13.3\% \\
DVC Pro (cheng2020) & -9.0\% & 7.2\% & -6.9\% & 17.2\% & -7.9\% & -4.1\% \\
HU\_ECCV20 & 2.4\% & 13.0\% & 10.8\% & -8.6\% & -5.4\% & -12.6\% \\
LU\_ECCV20 & 5.0\% & 8.4\% & 3.6\% & 11.7\% & 8.8\% & 8.4\% \\
Agustsson\_CVPR20 & & & & & -14.3\% & -16.9\% \\
NeRV\_NeurIPS21 & & & & & -19.1\% & -2.7\% \\
\textbf{Ours} & \textbf{-22.3\%} & \textbf{-6.0\%} & \textbf{-19.0\%} & \textbf{-24.3\%} & \textbf{-25.5\%} & \textbf{-18.2\%} \\
\bottomrule
\end{tabular}
\end{center}
\vskip -0.1in
\end{table*}

\subsection{Qualitative Results} \label{sec:exp-qualitative}

In Fig. \ref{fig:fig2}, we visualize the aleatoric and epistemic uncertainty in the first two frames of the BasketballDrill sequence, as well as the predictive uncertainty represented by the ensemble-based MV decoder. As we can see, the predictive uncertainty is more significant in regions where the motion cannot be accurately estimated or is too complicated to encode. Our uncertainty-aware model learns to effectively represent the underlying uncertainty with an ensemble of decoded MVs, and this uncertainty is retained until the final reconstruction.

\textcolor{revision-blue}{\textbf{Predictive uncertainty.} To demonstrate that the decoded representation from each member is indeed diversified to effectively capture the underlying uncertainty, we visualize the predictive uncertainty by modeling the ensemble predictions with a Gaussian mixture model and representing the predictive uncertainty with the variance term (see Eq.~\ref{eq:uncertainty}). For each 2D location, the variance would be larger if the predictions from the ensemble-based decoder are quite diversified, and the variance would be smaller if the predictions are consistent. We visualize the predictive uncertainty on three video sequences, BasketballDrive, RaceHorses, and Kimono1 in Fig.~\ref{fig:fig-5}. As we could see, the decoded representations from ensemble members are indeed diversified and larger predictive uncertainty corresponds to locations with large aleatoric uncertainty due to rapid motion and large epistemic uncertainty near object boundaries.}

\subsection{Ablation Study} \label{sec:exp-ablation}

\begin{figure}[!t]
\centering
\subfloat[]{\includegraphics[width=0.48\columnwidth]{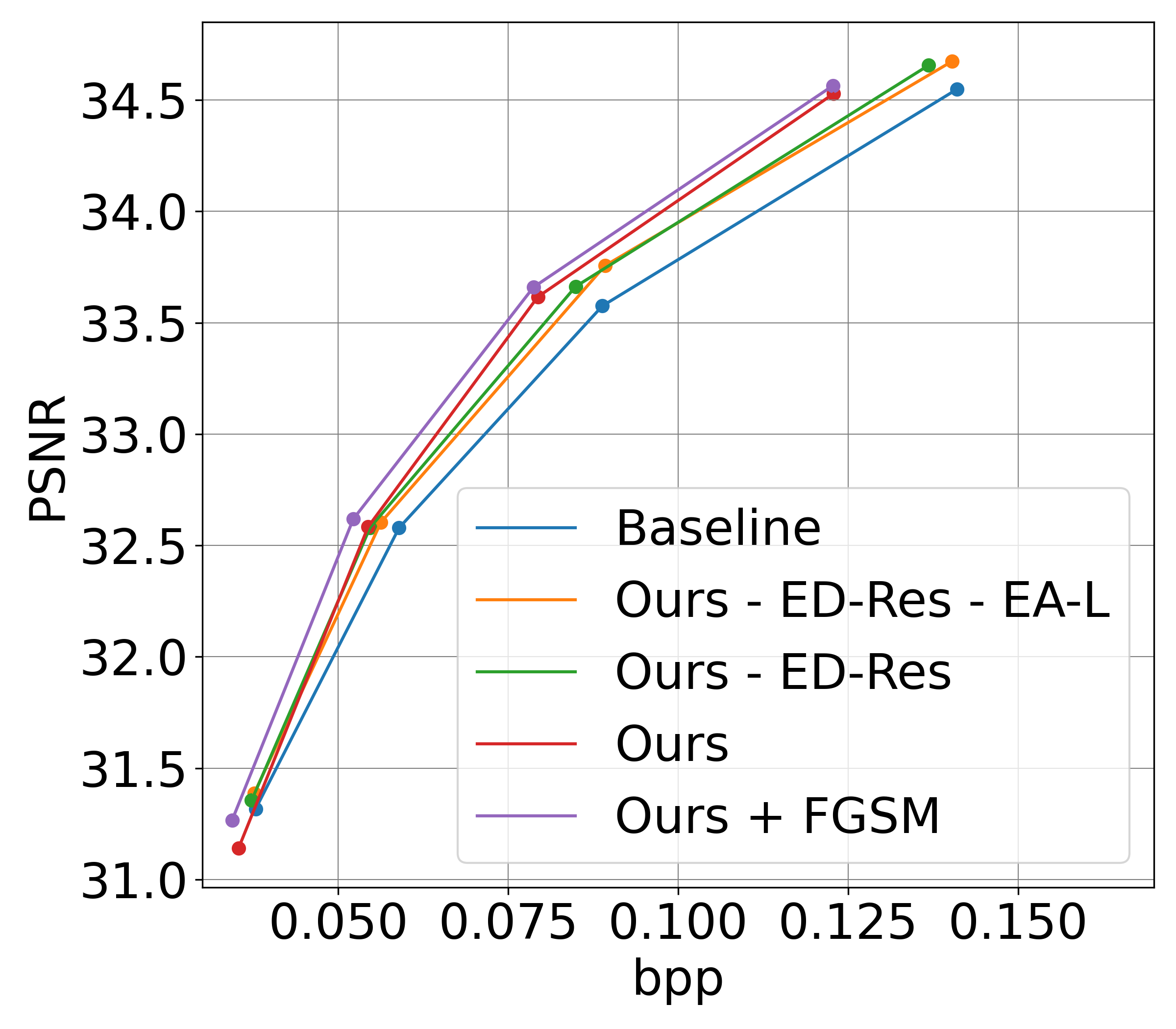}}
\label{fig:fig4-a}
\subfloat[]{\includegraphics[width=0.48\columnwidth]{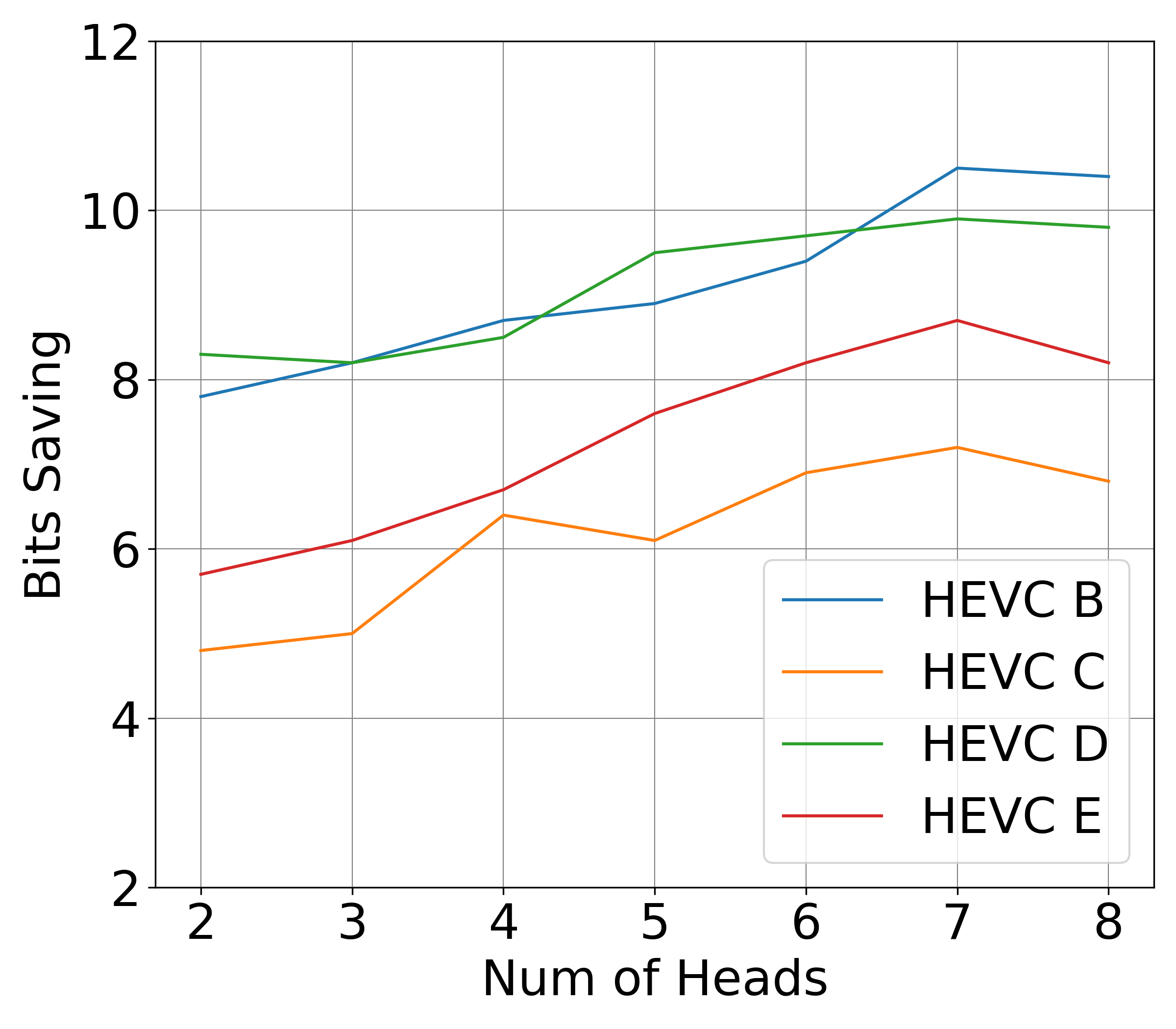}}
\label{fig:fig4-b}
\caption{(a) Effectiveness of various proposed modules. (b) Ablation study on the number of
    members in ensemble-based decoders.}
    \label{fig:fig4}
\end{figure}

\begin{figure}[!t]
\centering
\subfloat[]{\includegraphics[width=0.48\columnwidth]{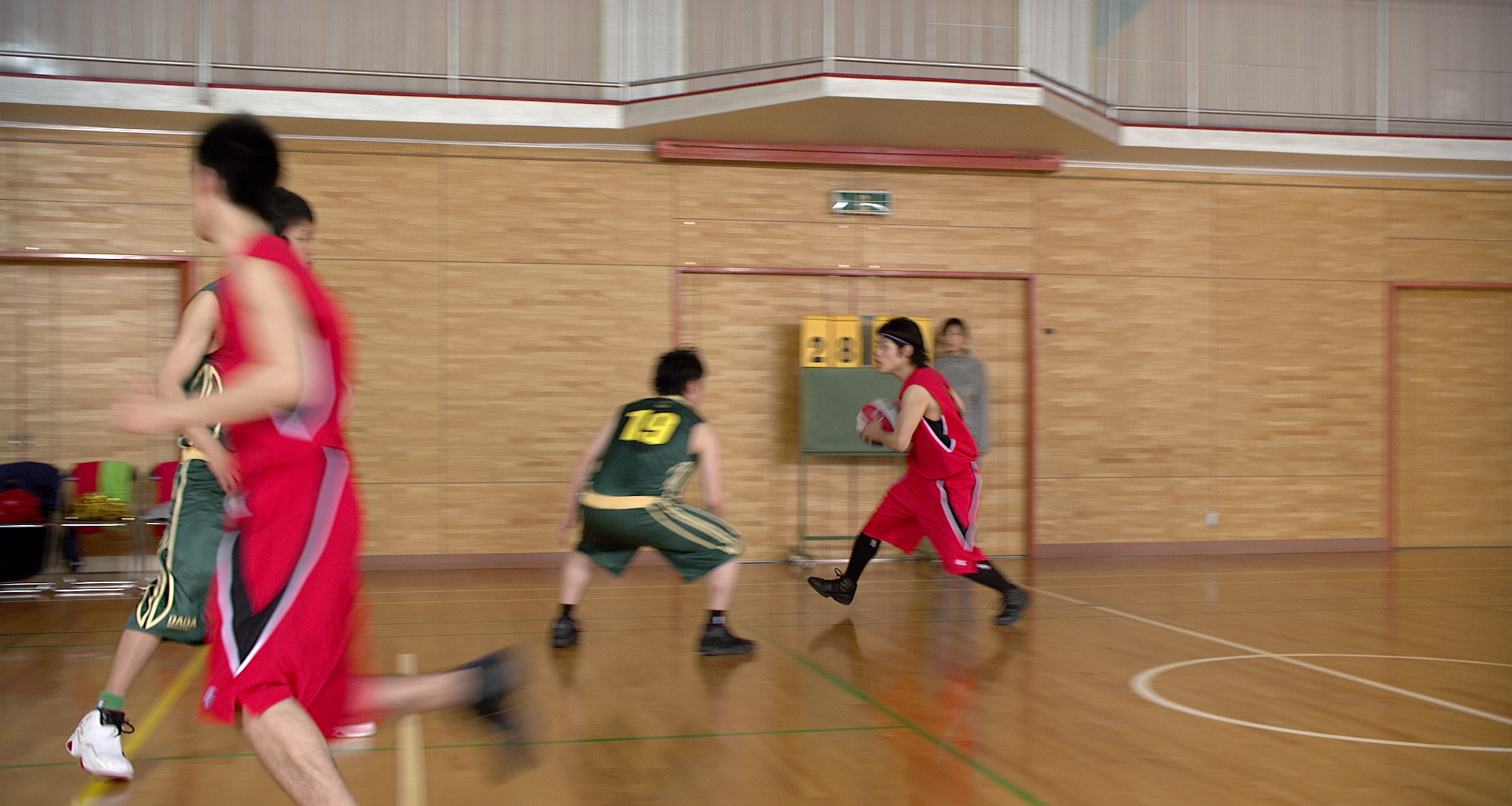}} \hfill
\subfloat[]{\includegraphics[width=0.48\columnwidth]{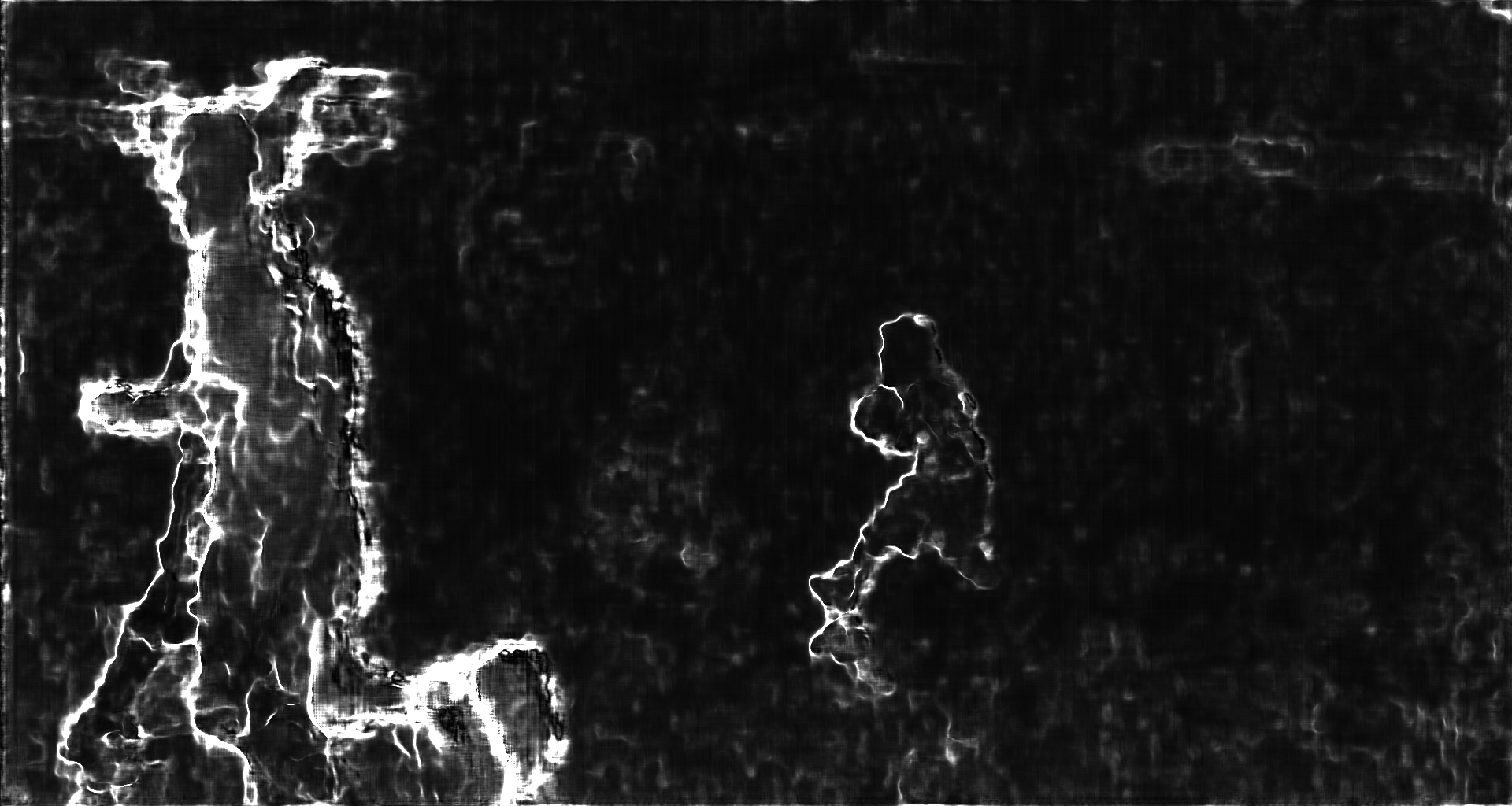}}

\subfloat[]{\includegraphics[width=0.48\columnwidth]{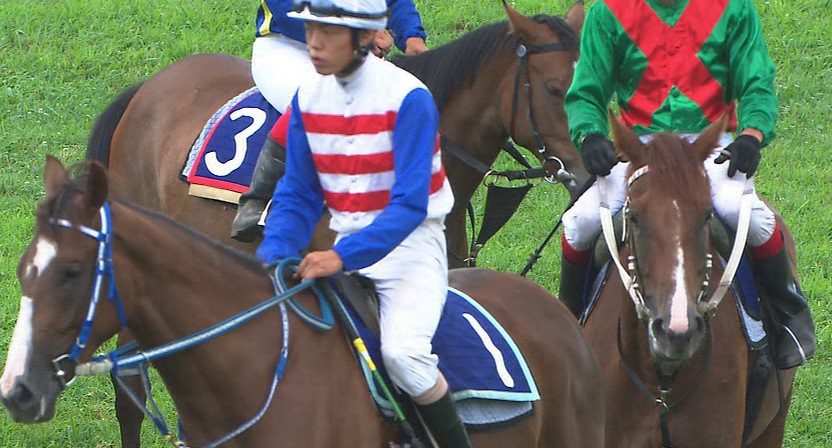}} \hfill
\subfloat[]{\includegraphics[width=0.48\columnwidth]{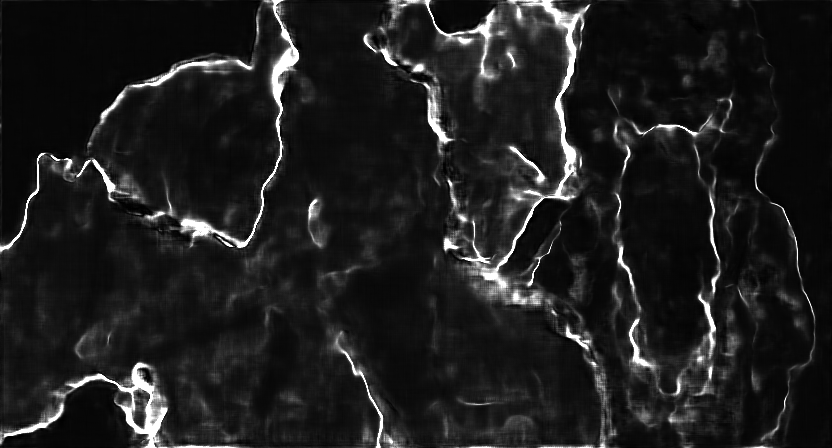}}

\subfloat[]{\includegraphics[width=0.48\columnwidth]{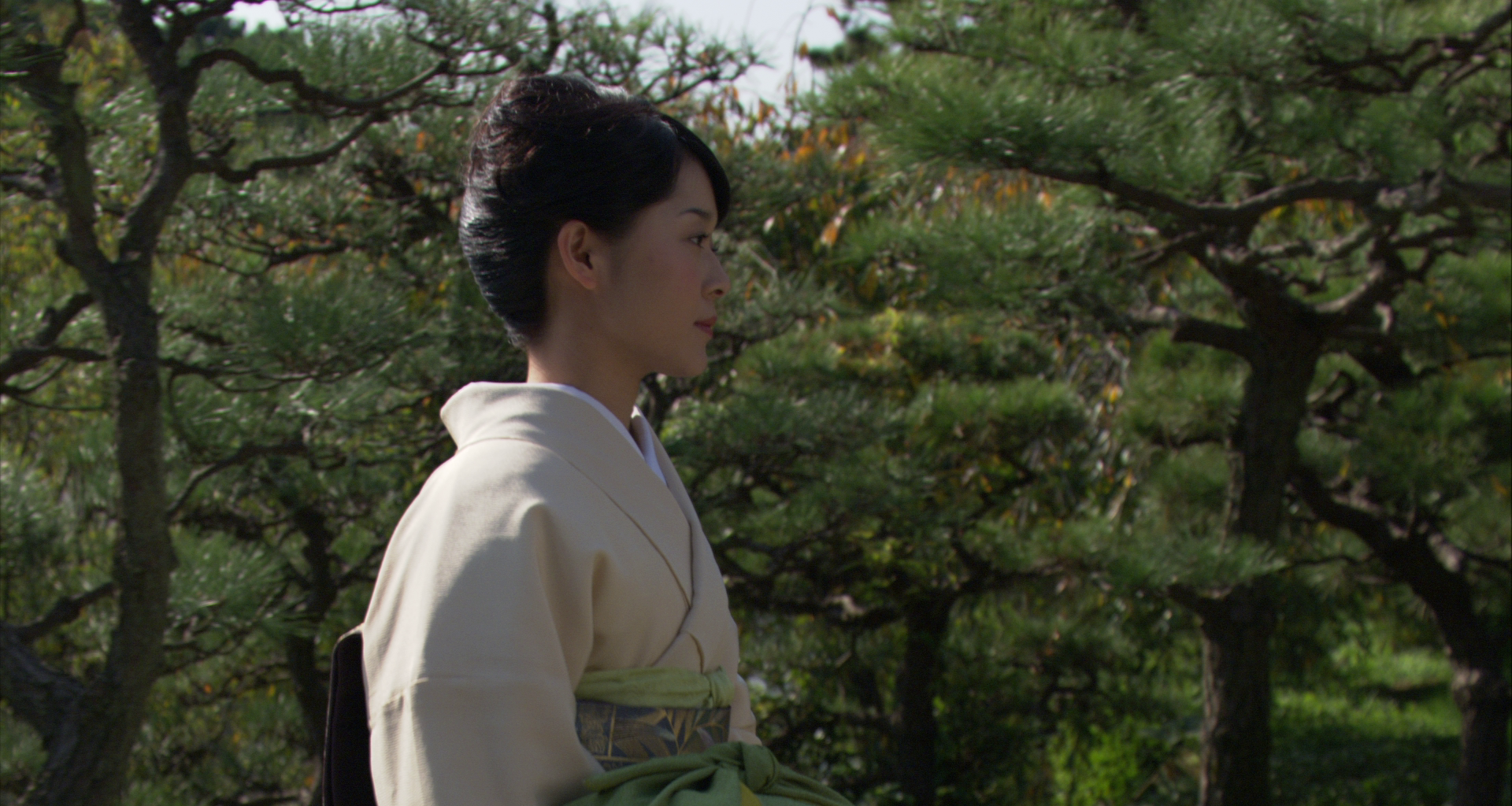}} \hfill
\subfloat[]{\includegraphics[width=0.48\columnwidth]{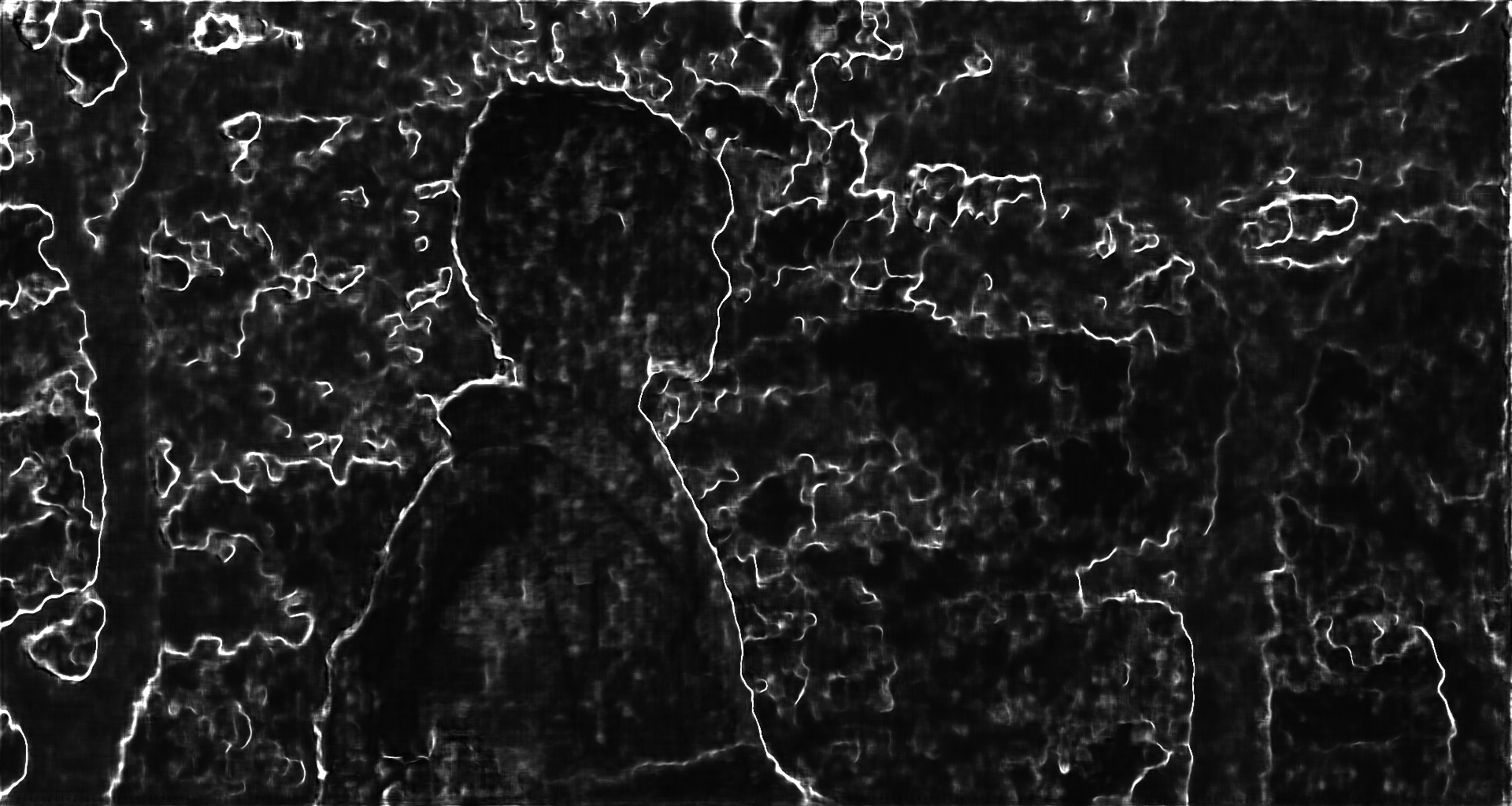}}
\caption{\textcolor{revision-blue}{Visualization of the predictive uncertainty represented by our proposed ensemble-based decoder on the first two frames in the BasketballDrive, RaceHorses, and Kimono1 sequence. The detailed calculations are presented in Eq.~\ref{eq:uncertainty}.}}
\label{fig:fig-5}
\end{figure}

\textbf{Effectiveness of various proposed modules.} We evaluate the effectiveness of ensemble-based decoders, ensemble-aware loss, and adversarial training with FGSM by running ablation experiments on the first 30 frames of all sequences in the HEVC dataset. We adopt the short training strategy for fast experimentation. The RD curves are presented in Fig. \ref{fig:fig4}(a) and the BD-rates are reported in Table \ref{tab:ablation-modules}. ``ED-MV'' and ``ED-Res'' represent ensemble-based decoders for MV and residual, and ``EA-L'' refers to training with ensemble-aware loss. Specifically, ``Ours'' is the baseline model augmented with ED-MV, ED-Res, and EA-L. The results show the efficacy of various proposed modules.

\begin{table*}[!t]
\caption{Ablation study on the effectiveness of each module. The performance is measured in BD rates using our baseline model as the anchor. \textbf{ED-MV}: ensemble-based MV decoder, \textbf{ED-Res}: ensemble-based residual decoder,  \textbf{EA-L}: ensemble-aware loss, \textbf{FGSM}: adversarial training with FGSM. Specifically, ``Ours'' is the baseline model augmented with ED-MV, ED-Res, and EA-L. All models have $h = 4$ considering the trade-off between performance and complexity.}
\label{tab:ablation-modules}
\begin{center}
\begin{tabular}{lcccc}
\toprule
\multicolumn{1}{c}{\bf Setting}  &\multicolumn{1}{c}{\bf HEVC B} &\multicolumn{1}{c}{\bf HEVC C} &\multicolumn{1}{c}{\bf HEVC D} &\multicolumn{1}{c}{\bf HEVC E} \\
\midrule
Ours - ED-MV - ED-Res - EA-L & 0.0 & 0.0 & 0.0 & 0.0 \\
Ours - ED-Res - EA-L & -5.8 & -3.1 & -4.0 & -3.7 \\
Ours - ED-Res & -7.0 & -4.0 & -6.7 & -4.8 \\
Ours & -8.7 & -6.4 & -8.5 & -6.7 \\
Ours + FGSM & -12.7 & -7.9 & -11.2 & -11.4 \\
\bottomrule
\end{tabular}
\end{center}
\end{table*}

\textbf{Ablation study on the number of ensemble members.} We investigate the model's performance with different numbers of members in the ensemble-based decoders on HEVC sequences. As shown in Fig. \ref{fig:fig4}(b), we train eight models with $h=1,\dots,8$ members where $h=1$ is the baseline without ensembling. We see that the ensemble-based decoder module is effective even with only two ensemble members, and the performance is improved with more ensemble members. Quantitative results are reported in Table \ref{tab:ablation-num-heads}.

\begin{table}[!t]
\caption{Ablation study on the number of ensemble members in the ensemble-based decoders. Performance is measured in BD rates using our baseline model as the anchor. We train eight models with $h = 1, \dots, 8$ using the fast training strategy.}
\label{tab:ablation-num-heads}
\vskip 0.15in
\begin{center}
\begin{tabular}{lcccc}
\toprule
\multicolumn{1}{c}{\bf Setting}  &\multicolumn{1}{c}{\bf HEVC B} &\multicolumn{1}{c}{\bf HEVC C} &\multicolumn{1}{c}{\bf HEVC D} &\multicolumn{1}{c}{\bf HEVC E} \\
\midrule
$h=1$ & 0.0 & 0.0 & 0.0 & 0.0 \\
$h=2$ & -7.8 & -4.8 & -8.3 & -5.7 \\
$h=3$ & -8.2 & -5.0 & -8.2 & -6.1 \\
$h=4$ & -8.7 & -6.4 & -8.5 & -6.7 \\
$h=5$ & -8.9 & -6.1 & -9.5 & -7.6 \\
$h=6$ & -9.4 & -6.9 & -9.7 & -8.2 \\
$h=7$ & -10.5 & -7.2 & -9.9 & -8.7 \\
$h=8$ & -10.4 & -6.8 & -9.8 & -8.2 \\
\bottomrule
\end{tabular}
\end{center}
\end{table}

\textbf{Ablation study on the ensemble-aware loss.} We introduced a novel ensemble-aware loss Eq. \ref{eq:ensemble-loss} in Section \ref{sec:method-ensemble} with a hyperparameter $k$, where the $k$ ensemble members with the smallest MSE are untouched, and the gradients of the other members are clipped.
Following the same experimental setting in the ablation study, we experiment with our model with $h=8$ ensemble members in the decoder for $k=2, 4, 6, 8$. The bits savings on HEVC B compared to the \textcolor{revision-blue}{model without ensemble-based decoders} are $-8.7\%$ $-10.4\%$, $-7.5\%$, and $-4.2\%$, respectively.

\textbf{Complexity analysis.} Most previous deep ensembles train multiple stand-alone models \cite{7298594,abbasi2017robustness,wang2020ensemble} or share very few shallow layers \cite{lee2015m} for the model to be effective. With an ensemble of 6 models, the inference complexity (in MACs) and model size (in the number of parameters) easily increase by $500\%$. In our ensemble-based decoder, the ensemble members share the backbone features, and we achieve superior results with a limited complexity increase. For one extra ensemble member in the MV and residual decoders, the complexity increases by $6\%$ and only $1\%$ in the model size. For our largest model with $h=8$, there is only a $48\%$ increase in complexity and $10\%$ in model size.

\textcolor{revision-blue}{\textbf{Choice of model designs.} Considering the trade-off between model performance and computational complexity, we chose $h=4$ and $k=1$ in our main experiments for an desirable performance with negligible complexity costs. For the implementation of FGSM, we followed a previous implementation of FGSM on ImageNet \cite{wong2020fast} and chose $\varepsilon = 4/255$.}

\section{Discussion}

In this paper, we studied the aleatoric and epistemic uncertainty in deep learning-based video compression and proposed to utilize an ensemble of intermediate predictions to represent the predictive uncertainty at decoding time. With ensemble-based decoders, our model can adequately model the uncertainties in the decoded MVs or residuals and effectively refine the motion compensation predictions and the reconstructed frames with the predictive uncertainty.

We investigated the performance of our uncertainty-aware decoding module and proposed a novel ensemble-aware loss to boost the diversity among the parallel ensemble branches in a single model. We also proposed to incorporate adversarial training for learning-based video codecs. Experimental results show the effectiveness of our approach.

Compared with one-stage learning-based video compression models, such as those based on 3D autoencoders \cite{pessoa2020end,habibian2019video}, two-stage motion compensation-based models can decode high-quality frames with low latency. However, intermediate predictions in these two-stage pipelines are not always accurate, and erroneous predictions could severely harm the performance of later stages, especially for out-of-distribution data. Therefore, it is critical to represent the predictive uncertainty, and our proposed ensemble-based decoder is a simple but very effective approach to capture such uncertainty. Future directions could involve modules on the encoder side to model and propagate the uncertainty to the decoders for an end-to-end uncertainty awareness.

\bibliography{example_paper}

\begin{thebibliography}{10}
\providecommand{\url}[1]{#1}
\csname url@samestyle\endcsname
\providecommand{\newblock}{\relax}
\providecommand{\bibinfo}[2]{#2}
\providecommand{\BIBentrySTDinterwordspacing}{\spaceskip=0pt\relax}
\providecommand{\BIBentryALTinterwordstretchfactor}{4}
\providecommand{\BIBentryALTinterwordspacing}{\spaceskip=\fontdimen2\font plus
\BIBentryALTinterwordstretchfactor\fontdimen3\font minus
  \fontdimen4\font\relax}
\providecommand{\BIBforeignlanguage}[2]{{%
\expandafter\ifx\csname l@#1\endcsname\relax
\typeout{** WARNING: IEEEtran.bst: No hyphenation pattern has been}%
\typeout{** loaded for the language `#1'. Using the pattern for}%
\typeout{** the default language instead.}%
\else
\language=\csname l@#1\endcsname
\fi
#2}}
\providecommand{\BIBdecl}{\relax}
\BIBdecl

\bibitem{cisco2020report}
\BIBentryALTinterwordspacing
Cisco, ``Cisco annual internet report (2018-2023) white paper,'' 2020.
  [Online]. Available:
  \url{https://www.cisco.com/c/en/us/solutions/collateral/executive-perspectives/annual-internet-report/white-paper-c11-741490.html}
\BIBentrySTDinterwordspacing

\bibitem{rippel2019learned}
O.~Rippel, S.~Nair, C.~Lew, S.~Branson, A.~G. Anderson, and L.~Bourdev,
  ``Learned video compression,'' in \emph{Proceedings of the IEEE/CVF
  International Conference on Computer Vision}, 2019, pp. 3454--3463.

\bibitem{9072487}
G.~Lu, X.~Zhang, W.~Ouyang, L.~Chen, Z.~Gao, and D.~Xu, ``An end-to-end
  learning framework for video compression,'' \emph{IEEE Transactions on
  Pattern Analysis and Machine Intelligence}, pp. 1--1, 2020.

\bibitem{Agustsson_2020_CVPR}
E.~Agustsson, D.~Minnen, N.~Johnston, J.~Balle, S.~J. Hwang, and G.~Toderici,
  ``Scale-space flow for end-to-end optimized video compression,'' in
  \emph{Proceedings of the IEEE/CVF Conference on Computer Vision and Pattern
  Recognition (CVPR)}, June 2020.

\bibitem{hao2021nerv}
H.~Chen, B.~He, H.~Wang, Y.~Ren, S.-N. Lim, and A.~Shrivastava, ``Nerv: Neural
  representations for videos s,'' in \emph{NeurIPS}, 2021.

\bibitem{tomar2006converting}
S.~Tomar, ``Converting video formats with ffmpeg,'' \emph{Linux Journal}, vol.
  2006, no. 146, p.~10, 2006.

\bibitem{der2009aleatory}
A.~Der~Kiureghian and O.~Ditlevsen, ``Aleatory or epistemic? does it matter?''
  \emph{Structural safety}, vol.~31, no.~2, pp. 105--112, 2009.

\bibitem{kendall2017uncertainties}
A.~Kendall and Y.~Gal, ``What uncertainties do we need in bayesian deep
  learning for computer vision?'' \emph{arXiv preprint arXiv:1703.04977}, 2017.

\bibitem{Gal2016Uncertainty}
Y.~Gal, ``Uncertainty in deep learning,'' Ph.D. dissertation, University of
  Cambridge, 2016.

\bibitem{NIPS2017_9ef2ed4b}
B.~Lakshminarayanan, A.~Pritzel, and C.~Blundell, ``Simple and scalable
  predictive uncertainty estimation using deep ensembles,'' in \emph{Advances
  in Neural Information Processing Systems}, I.~Guyon, U.~V. Luxburg,
  S.~Bengio, H.~Wallach, R.~Fergus, S.~Vishwanathan, and R.~Garnett, Eds.,
  vol.~30.\hskip 1em plus 0.5em minus 0.4em\relax Curran Associates, Inc.,
  2017.

\bibitem{mackay1992practical}
D.~J. MacKay, ``A practical bayesian framework for backpropagation networks,''
  \emph{Neural computation}, vol.~4, no.~3, pp. 448--472, 1992.

\bibitem{Hinton1995BayesianLF}
G.~E. Hinton and R.~Neal, ``Bayesian learning for neural networks,'' 1995.

\bibitem{gal2016dropout}
Y.~Gal and Z.~Ghahramani, ``Dropout as a bayesian approximation: Representing
  model uncertainty in deep learning,'' in \emph{international conference on
  machine learning}.\hskip 1em plus 0.5em minus 0.4em\relax PMLR, 2016, pp.
  1050--1059.

\bibitem{374138}
D.~Nix and A.~Weigend, ``Estimating the mean and variance of the target
  probability distribution,'' in \emph{Proceedings of 1994 IEEE International
  Conference on Neural Networks (ICNN'94)}, vol.~1, 1994, pp. 55--60 vol.1.

\bibitem{goodfellow2014explaining}
I.~J. Goodfellow, J.~Shlens, and C.~Szegedy, ``Explaining and harnessing
  adversarial examples,'' \emph{arXiv preprint arXiv:1412.6572}, 2014.

\bibitem{pessoa2020end}
J.~Pessoa, H.~Aidos, P.~Tom{\'a}s, and M.~A. Figueiredo, ``End-to-end learning
  of video compression using spatio-temporal autoencoders,'' in \emph{2020 IEEE
  Workshop on Signal Processing Systems (SiPS)}.\hskip 1em plus 0.5em minus
  0.4em\relax IEEE, 2020, pp. 1--6.

\bibitem{habibian2019video}
A.~Habibian, T.~v. Rozendaal, J.~M. Tomczak, and T.~S. Cohen, ``Video
  compression with rate-distortion autoencoders,'' in \emph{Proceedings of the
  IEEE/CVF International Conference on Computer Vision}, 2019, pp. 7033--7042.

\bibitem{lu2019dvc}
G.~Lu, W.~Ouyang, D.~Xu, X.~Zhang, C.~Cai, and Z.~Gao, ``Dvc: An end-to-end
  deep video compression framework,'' in \emph{Proceedings of the IEEE/CVF
  Conference on Computer Vision and Pattern Recognition}, 2019, pp.
  11\,006--11\,015.

\bibitem{Ranjan_2017_CVPR}
A.~Ranjan and M.~J. Black, ``Optical flow estimation using a spatial pyramid
  network,'' in \emph{Proceedings of the IEEE Conference on Computer Vision and
  Pattern Recognition (CVPR)}, July 2017.

\bibitem{9185043}
P.~He, H.~Li, H.~Wang, S.~Wang, X.~Jiang, and R.~Zhang, ``Frame-wise detection
  of double hevc compression by learning deep spatio-temporal representations
  in compression domain,'' \emph{IEEE Transactions on Multimedia}, vol.~23, pp.
  3179--3192, 2021.

\bibitem{8447515}
F.~Luo, S.~Wang, S.~Wang, X.~Zhang, S.~Ma, and W.~Gao, ``Gpu-based hierarchical
  motion estimation for high efficiency video coding,'' \emph{IEEE Transactions
  on Multimedia}, vol.~21, no.~4, pp. 851--862, 2019.

\bibitem{9681152}
M.~Lu, T.~Chen, Z.~Dai, D.~Wang, D.~Ding, and Z.~Ma, ``Decoder-side cross
  resolution synthesis for video compression enhancement,'' \emph{IEEE
  Transactions on Multimedia}, pp. 1--1, 2022.

\bibitem{7736114}
S.~Wang, X.~Zhang, X.~Liu, J.~Zhang, S.~Ma, and W.~Gao, ``Utility-driven
  adaptive preprocessing for screen content video compression,'' \emph{IEEE
  Transactions on Multimedia}, vol.~19, no.~3, pp. 660--667, 2017.

\bibitem{hendrycks2016baseline}
D.~Hendrycks and K.~Gimpel, ``A baseline for detecting misclassified and
  out-of-distribution examples in neural networks,'' \emph{arXiv preprint
  arXiv:1610.02136}, 2016.

\bibitem{abbasi2017robustness}
M.~Abbasi and C.~Gagn{\'e}, ``Robustness to adversarial examples through an
  ensemble of specialists,'' \emph{arXiv preprint arXiv:1702.06856}, 2017.

\bibitem{58871}
L.~Hansen and P.~Salamon, ``Neural network ensembles,'' \emph{IEEE Transactions
  on Pattern Analysis and Machine Intelligence}, vol.~12, no.~10, pp.
  993--1001, 1990.

\bibitem{WOLPERT1992241}
D.~H. Wolpert, ``Stacked generalization,'' \emph{Neural Networks}, vol.~5,
  no.~2, pp. 241--259, 1992.

\bibitem{Perrone1993}
M.~Perrone and L.~Cooper, ``When networks disagree: Ensemble methods for hybrid
  neural networks,'' \emph{Neural networks for speech and image processing}, 08
  1993.

\bibitem{NIPS1994_b8c37e33}
A.~Krogh and J.~Vedelsby, ``Neural network ensembles, cross validation, and
  active learning,'' in \emph{Advances in Neural Information Processing
  Systems}, G.~Tesauro, D.~Touretzky, and T.~Leen, Eds., vol.~7.\hskip 1em plus
  0.5em minus 0.4em\relax MIT Press, 1995.

\bibitem{lee2015m}
S.~Lee, S.~Purushwalkam, M.~Cogswell, D.~Crandall, and D.~Batra, ``Why m heads
  are better than one: Training a diverse ensemble of deep networks,'' 2015.

\bibitem{7298594}
C.~Szegedy, W.~Liu, Y.~Jia, P.~Sermanet, S.~Reed, D.~Anguelov, D.~Erhan,
  V.~Vanhoucke, and A.~Rabinovich, ``Going deeper with convolutions,'' in
  \emph{2015 IEEE Conference on Computer Vision and Pattern Recognition
  (CVPR)}, 2015, pp. 1--9.

\bibitem{NEURIPS2018_be3087e7}
T.~Garipov, P.~Izmailov, D.~Podoprikhin, D.~P. Vetrov, and A.~G. Wilson, ``Loss
  surfaces, mode connectivity, and fast ensembling of dnns,'' in \emph{Advances
  in Neural Information Processing Systems}, S.~Bengio, H.~Wallach,
  H.~Larochelle, K.~Grauman, N.~Cesa-Bianchi, and R.~Garnett, Eds.,
  vol.~31.\hskip 1em plus 0.5em minus 0.4em\relax Curran Associates, Inc.,
  2018.

\bibitem{fort2020deep}
S.~Fort, H.~Hu, and B.~Lakshminarayanan, ``Deep ensembles: A loss landscape
  perspective,'' 2020.

\bibitem{lu2020content}
G.~Lu, C.~Cai, X.~Zhang, L.~Chen, W.~Ouyang, D.~Xu, and Z.~Gao, ``Content
  adaptive and error propagation aware deep video compression,'' in
  \emph{European Conference on Computer Vision}.\hskip 1em plus 0.5em minus
  0.4em\relax Springer, 2020, pp. 456--472.

\bibitem{hu2020improving}
Z.~Hu, Z.~Chen, D.~Xu, G.~Lu, W.~Ouyang, and S.~Gu, ``Improving deep video
  compression by resolution-adaptive flow coding,'' in \emph{European
  Conference on Computer Vision}.\hskip 1em plus 0.5em minus 0.4em\relax
  Springer, 2020, pp. 193--209.

\bibitem{wang2020ensemble}
Y.~Wang, D.~Liu, S.~Ma, F.~Wu, and W.~Gao, ``Ensemble learning-based
  rate-distortion optimization for end-to-end image compression,'' \emph{IEEE
  Transactions on Circuits and Systems for Video Technology}, 2020.

\bibitem{fort2019deep}
S.~Fort, H.~Hu, and B.~Lakshminarayanan, ``Deep ensembles: A loss landscape
  perspective,'' \emph{arXiv preprint arXiv:1912.02757}, 2019.

\bibitem{bartlett1998boosting}
P.~Bartlett, Y.~Freund, W.~S. Lee, and R.~E. Schapire, ``Boosting the margin: A
  new explanation for the effectiveness of voting methods,'' \emph{The annals
  of statistics}, vol.~26, no.~5, pp. 1651--1686, 1998.

\bibitem{szegedy2013intriguing}
C.~Szegedy, W.~Zaremba, I.~Sutskever, J.~Bruna, D.~Erhan, I.~Goodfellow, and
  R.~Fergus, ``Intriguing properties of neural networks,'' \emph{arXiv preprint
  arXiv:1312.6199}, 2013.

\bibitem{balle2016end}
J.~Ball{\'e}, V.~Laparra, and E.~P. Simoncelli, ``End-to-end optimization of
  nonlinear transform codes for perceptual quality,'' in \emph{2016 Picture
  Coding Symposium (PCS)}.\hskip 1em plus 0.5em minus 0.4em\relax IEEE, 2016,
  pp. 1--5.

\bibitem{xue2019video}
T.~Xue, B.~Chen, J.~Wu, D.~Wei, and W.~T. Freeman, ``Video enhancement with
  task-oriented flow,'' \emph{International Journal of Computer Vision}, vol.
  127, no.~8, pp. 1106--1125, 2019.

\bibitem{wang2003multiscale}
Z.~Wang, E.~P. Simoncelli, and A.~C. Bovik, ``Multiscale structural similarity
  for image quality assessment,'' in \emph{The Thrity-Seventh Asilomar
  Conference on Signals, Systems \& Computers, 2003}, vol.~2.\hskip 1em plus
  0.5em minus 0.4em\relax Ieee, 2003, pp. 1398--1402.

\bibitem{loshchilov2017decoupled}
I.~Loshchilov and F.~Hutter, ``Decoupled weight decay regularization,''
  \emph{arXiv preprint arXiv:1711.05101}, 2017.

\bibitem{sullivan2012overview}
G.~J. Sullivan, J.-R. Ohm, W.-J. Han, and T.~Wiegand, ``Overview of the high
  efficiency video coding (hevc) standard,'' \emph{IEEE Transactions on
  circuits and systems for video technology}, vol.~22, no.~12, pp. 1649--1668,
  2012.

\bibitem{mercat2020uvg}
A.~Mercat, M.~Viitanen, and J.~Vanne, ``Uvg dataset: 50/120fps 4k sequences for
  video codec analysis and development,'' in \emph{Proceedings of the 11th ACM
  Multimedia Systems Conference}, 2020, pp. 297--302.

\bibitem{wang2016mcl}
H.~Wang, W.~Gan, S.~Hu, J.~Y. Lin, L.~Jin, L.~Song, P.~Wang, I.~Katsavounidis,
  A.~Aaron, and C.-C.~J. Kuo, ``Mcl-jcv: a jnd-based h. 264/avc video quality
  assessment dataset,'' in \emph{2016 IEEE International Conference on Image
  Processing (ICIP)}.\hskip 1em plus 0.5em minus 0.4em\relax IEEE, 2016, pp.
  1509--1513.

\bibitem{Cheng_2020_CVPR}
Z.~Cheng, H.~Sun, M.~Takeuchi, and J.~Katto, ``Learned image compression with
  discretized gaussian mixture likelihoods and attention modules,'' in
  \emph{Proceedings of the IEEE/CVF Conference on Computer Vision and Pattern
  Recognition (CVPR)}, June 2020.

\bibitem{Bjntegaard2001CalculationOA}
G.~Bj{\o}ntegaard, ``Calculation of average psnr differences between
  rd-curves,'' 2001.

\bibitem{wong2020fast}
E.~Wong, L.~Rice, and J.~Z. Kolter, ``Fast is better than free: Revisiting
  adversarial training,'' \emph{arXiv preprint arXiv:2001.03994}, 2020.

\end{thebibliography}
\bibliographystyle{IEEEtran}

\vspace{.2in}

\begin{wrapfigure}{l}{25mm} 
\includegraphics[width=1in,height=1.25in,keepaspectratio]{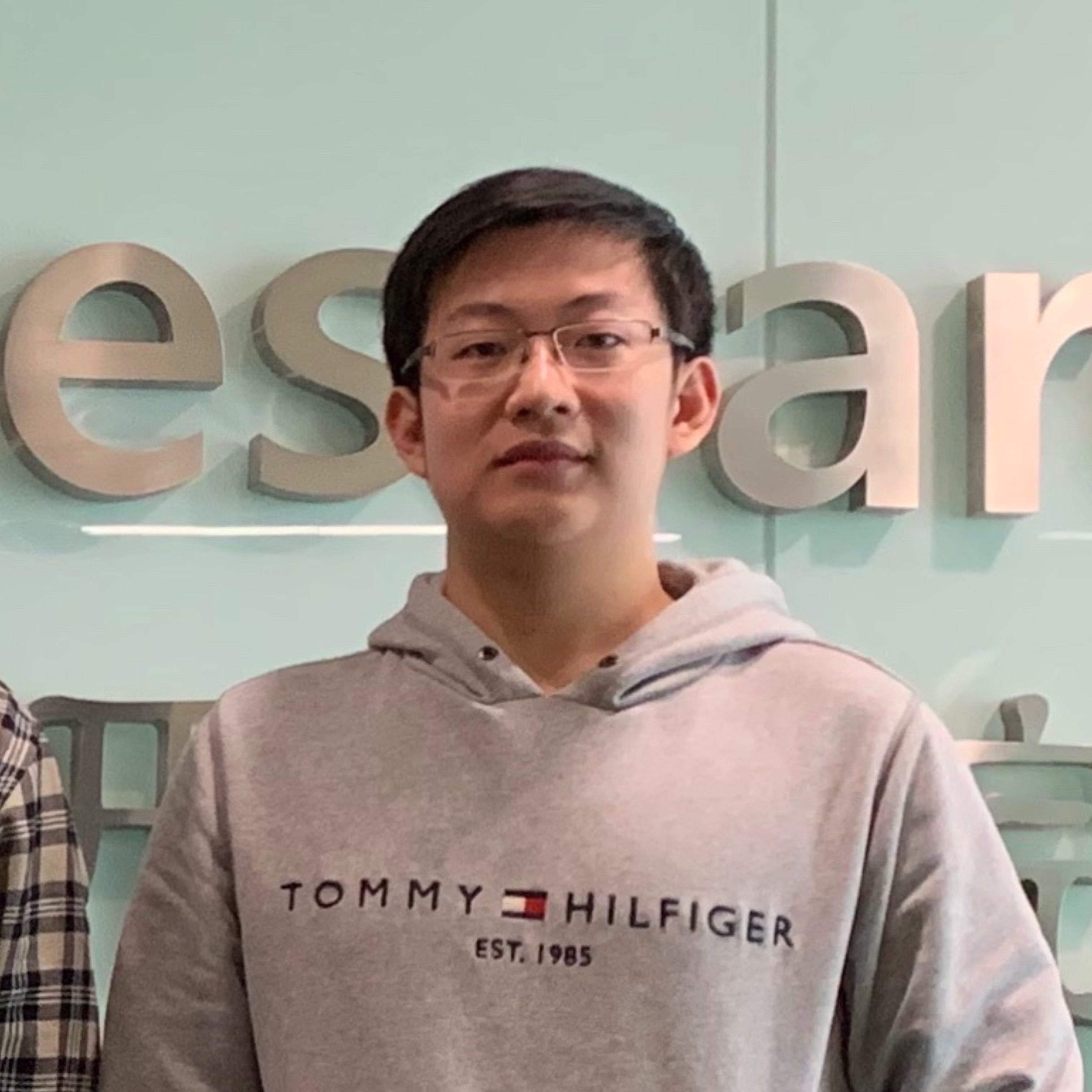}
\end{wrapfigure}\par
\textbf{Wufei Ma} is a Ph.D. student in Computer Science at Johns Hopkins University. He obtained his B.S. degree with \textit{summa cum laude} honor in Computer Science and Mathematics from Rensselaer Polytechnic Institute in 2020. His research primarily focuses on representation learning and 3D vision.

\vspace{.2in}

\begin{wrapfigure}{l}{25mm} 
\includegraphics[width=1in,height=1.25in,keepaspectratio]{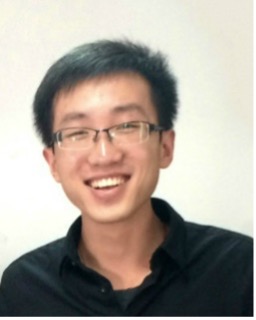}
\end{wrapfigure}\par
\textbf{Jiahao Li} received the B.S. degree in computer science and technology from the Harbin Institute of Technology in 2014, and the Ph.D. degree from Peking University in 2019. He is currently a Senior Researcher with the Media Computing Group, Microsoft Research Asia. His research interests include neural video compression and other video tasks, like video backbone design and video representation learning. He has more than thirty published papers, standard proposals, and patents in the related area.

\vspace{.2in}

\begin{wrapfigure}{l}{25mm} 
\includegraphics[width=1in,height=1.25in,keepaspectratio]{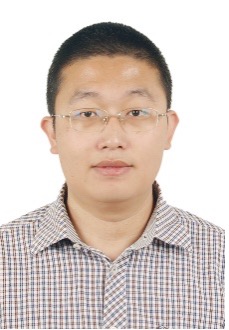}
\end{wrapfigure}\par
\textbf{Bin Li} received the B.S. and Ph.D. degrees in electronic engineering from the University of Science and Technology of China (USTC), Hefei, Anhui, China, in 2008 and 2013, respectively. He joined Microsoft Research Asia (MSRA), Beijing, China, in 2013 and now he is a Principal Researcher. He has authored or co-authored over 50 papers. He holds over 30 granted or pending U.S. patents in the area of image and video coding. He has more than 40 technical proposals that have been adopted by Joint Collaborative Team on Video Coding. His current research interests include video coding, processing, transmission, and communication. Dr. Li received the best paper award for the International Conference on Mobile and Ubiquitous Multimedia from Association for Computing Machinery in 2011. He received the Top 10\% Paper Award of 2014 IEEE International Conference on Image Processing. He received the best paper award of IEEE Visual Communications and Image Processing 2017.

\vspace{.2in}

\begin{wrapfigure}{l}{25mm} 
\includegraphics[width=1in,height=1.25in,keepaspectratio]{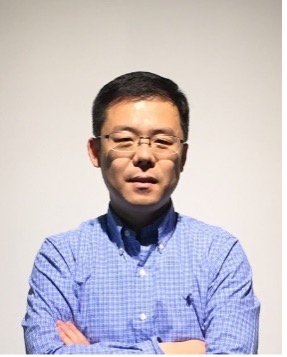}
\end{wrapfigure}\par
\textbf{Yan Lu} received his Ph.D. degree in computer science from Harbin Institute of Technology, China. He joined Microsoft Research Asia in 2004, where he is now a Partner Research Manager and manages research on media computing and communication. He and his team have transferred many key technologies and research prototypes to Microsoft products. From 2001 to 2004, he was a team lead of video coding group in the JDL Lab, Institute of Computing Technology, China. From 1999 to 2000, he was with the City University of Hong Kong as a research assistant. Yan Lu has broad research interests in the fields of real-time communication, computer vision, video analytics, audio enhancement, virtualization, and mobile-cloud computing. He holds 30+ granted US patents and has published 100+ papers in refereed journals and conference proceedings.

\vfill

\end{document}